\documentclass[lettersize,journal]{IEEEtran}

\usepackage{amsmath,amsfonts}
\usepackage{algorithmic}
\usepackage{algorithm}
\usepackage{array}
\usepackage[caption=false,font=footnotesize,labelfont=sf, textfont=sf]{subfig}
\usepackage{textcomp}
\usepackage{stfloats}
\usepackage{url}
\usepackage{verbatim}
\usepackage{graphicx}
\usepackage{cite}
\usepackage{bm}
\usepackage{amssymb}
\usepackage{multirow}
\makeatletter
\let\NAT@parse\undefined
\makeatother
\usepackage{hyperref} 
\begin{document}
% \bstctlcite{IEEEexample:BSTcontrol}
\title{GT-CausIn: a novel causal-based\\insight for traffic prediction}

\author{Ting~Gao,~Rodrigo~Kappes~Marques,~Lei Yu}

\markboth{Journal of \LaTeX\ Class Files}%
% \markboth{IEEE TRANSACTIONS ON INTELLIGENT TRANSPORTATION SYSTEMS}
{Shell \MakeLowercase{\textit{et al.}}: A Sample Article Using IEEEtran.cls for IEEE Journals}

\IEEEpubid{0000--0000/00\$00.00~\copyright~2021 IEEE}

\maketitle

\begin{abstract}
Traffic forecasting is an important application of spatiotemporal series prediction. Among different methods, graph neural networks have achieved so far the most promising results, learning relations between graph nodes then becomes a crucial task. However, improvement space is very limited when these relations are learned in a node-to-node manner. The challenge stems from (1) obscure temporal dependencies between different stations, (2) difficulties in defining variables beyond the node level, and (3) no ready-made method to validate the learned relations. To confront these challenges, we define legitimate traffic causal variables to discover the causal relation inside the traffic network, which is carefully checked with statistic tools and case analysis. We then present a novel model named \textit{Graph Spatial-Temporal Network Based on Causal Insight} (GT-CausIn), where prior learned causal information is integrated with graph diffusion layers and temporal convolutional network (TCN) layers. Experiments are carried out on two real-world traffic datasets: PEMS-BAY and METR-LA, which show that GT-CausIn significantly outperforms the state-of-the-art models on mid-term and long-term prediction.
\end{abstract}


\section{Introduction}

\IEEEPARstart{A}{s} an important aspect of smart cities\cite{NAGY2018148}, traffic forecasting is gaining more and more popularity in this data era. Traffic forecasting aims at predicting future traffic volumes with historical records. Different models have been proposed and evolved. Knowledge-driven models focus on defining functional characteristics of traffic elements and simulating various traffic service demands \cite{cascetta2013transportation}, where analysis is generally theory-based and requires constrained mathematical assumptions. Data-driven models contain three major categories: time-series analysis methods (e.g., ARIMA \cite{hamed1995short}, SARIMA \cite{williams2003modeling}), traditional machine learning methods (e.g., SVM \cite{smola2004tutorial}, SVR \cite{evgeniou2000regularization}) and deep learning based methods (e.g., CNN \cite{ma2017learning, guo2019deep}).  

Deep learning methods outperform the others by a large margin and have become the mainstream solution in the industry. Guo \textit{et al.} \cite{guo2019deep} partition a city into a grid-based map and use 3D convolution to capture the spatiotemporal correlation of traffic data.  Geng \textit{et al.} \cite{geng2019spatiotemporal} build road graphs from multiple views based on pixel images and design a Contextual Gated Recurrent Neural Network (CGRNN) for information integration. Li \textit{et al.} \cite{li2017diffusion} model traffic flow as a diffusion process for directed graph, encoder-decoder structure and scheduled sampling are further proposed to reduce error accumulation and improve model performance. Zheng \textit{et al.} \cite{zheng2020gman} skip convolution layers and benefit from multi-head attention layers to embed spatial and temporal information.
 
The traffic road network provides a natural premium connection map for traffic flow analysis, nevertheless, we argue that its inherent regime is not limited to node-to-node impacts. For example, the inflow of a highway service station is equal to its outflow within a certain period of time. Therefore, it is interesting to analyze the relations between different traffic variables. In recent years, causal discovery \cite{pearl2009causality} opens up the door to the analysis of causal relations behind statistical correlations and is widely explored in the computational biology domain \cite{pranay2021causal, cundy2021bcd}. However, to the best of our knowledge, few works consider causal observation for spatiotemporal forecasting. 

In this work, we focus on improving traffic speed prediction with insight revealed by a causal discovery program. Speed change is thoroughly used for causal discovery, as it is a good indication of traffic conditions and can lead to changes in neighboring roads. 
In summary, our main contributions are: 
\begin{itemize}
    \item {We define legitimate traffic variables for causal discovery and conduct dense experiments on the PEMS-BAY dataset to learn their causal relations. We also find out that this knowledge can be generalized to the METR-LA dataset.}
    \item {We present an effective and efficient framework to capture spatial-temporal dependencies, which is named \textit{Graph Spatial-Temporal Network Based on Causal Insight} (GT-CausIn). The core idea is to assemble causal insight, spatial dependency modeling, and temporal dependency modeling in a way that information can flexibly flow between different perspectives at different scales.}
    \item {We validate our model on two real-world public datasets and achieve state-of-the-art performance.} 
\end{itemize}

\section{Related work}

{\bf{Causal structural discovery}}   The objective of causal structural discovery is to combine statistical and logical reasoning to obtain causal relations between variables, which are represented by a graph. Fast Causal Inference (FCI) was proposed in \cite{spirtes2000constructing}, which estimates the Markov equivalence class of causal Maximal Ancestral Graph (MAG). 
\IEEEpubidadjcol
Recently, many papers came up with ideas to improve the high computational cost of causal inference. Zečević \textit{et al.} \cite{zevcevic2021interventional} step towards this problem by learning interventional distributions with an over-parameterized sum-product network. Akbari \textit{et al.} \cite{akbari2021recursive} propose a recursive constraint-based method that removes a specific type of variables at each iteration. Rohekar \textit{et al.} \cite{rohekar2021iterative} further introduce Iterative Causal Discovery (ICD) based on FCI. Each iteration of ICD refines a graph recovered from previous iterations with a smaller set of conditions and higher statistical reliability. As a result, ICD requires fewer conditional independence tests and is more computational resources friendly. In this work, we benefit from ICD to discover the causal structure of defined variables.

{\bf{Spatial dependency}}    Because of its intrinsic advantages of exploring topological spaces, graph-based neural networks stand out from diverse deep learning models and achieve so far the best results. Cao \textit{et al.} \cite{cao2020spectral} develop the StemGNN layer to extract in-series and intra-series correlations with the help of Graph Fourier Transformer (GFT) and Inverse Graph Fourier Transformer (IGFT). Derrow-Pinion \textit{et al.} \cite{derrow2021eta} rethink the relations between road segments and collects information from node level, segment level, and super-segment level. Dynamic transition matrices are adopted in many works \cite{du2020traffic, han2021dynamic, shin2022pgcn}, where nodes with similar embeddings are assumed to have higher connection weights. However, there may be some problems with this intuitive hypothesis. Matrices are fixed after training, and embedding similarity may yaw because of sensor dysfunction and extreme external situations (e.g., hot/cold weather), as a result, the adaptive matrix can get biased. In this work, we employ graph diffusion layers designed in \cite{li2017diffusion} with predefined roads as a transition matrix, which provides reliable physical connections in the real world.

{\bf{Temporal dependency}}    The most popular way to discover temporal dependency is through RNN-based networks. Long Short-Term Memory neural network (LSTM) is adopted in the work \cite{lin2018predicting, yao2018deep}, while Li \textit{et al.} \cite{li2017diffusion} and Jin \textit{et al.} \cite{jin2022gan} arrange RNN block in their core model structure. Geng \textit{et al.} \cite{geng2019spatiotemporal} and Lv \textit{et al.} \cite{lv2020temporal} apply Gated Recurrent Unit (GRU). Due to the time-consuming problem and complex gate mechanism of RNN-based networks, we assign TCN to model temporal dependency, which captures long sequences in a non-recursive manner \cite{han2021dynamic, shin2022pgcn}.

\section{Preparatory work}
\subsection{Problem formulation}
In a traffic forecasting scenario, sensors are scattered on the road to record traffic speed at a given time, and thus can be represented as a directed graph. In this work, a graph is denoted by $\mathcal{G}(\mathcal{V}, \mathcal{E}, \mathbf{W})$ where $\mathcal{V}$ is the node set ($|\mathcal{V}|=N$), $\mathcal{E}$ is the edge set and $\mathbf{W}$ is the adjacency matrix ($|\mathbf{W}|=N\times N$). Main notations used in the paper are summarised in Appendix \ref{notation}. In the real world, the adjacency matrix is a representation of proximity and connectivity between node sensors.
 
Suppose each node's feature is of dimension $P$ (e.g., velocity, volume, occupancy), at timestamps $t$, a graph signal can be defined as $\mathbf{X}^{(t)}(\mathbf{X}^{t}\in\mathbb{R}^{N\times P})$. The predicted signal at time $t$ is noted as $\hat{\mathbf{X}}^{(t)}(\hat{\mathbf{X}}^{(t)}\in\mathbb{R}^{N\times Q})$, where $Q$ is the number of the desired features. We consider the prediction task as using $T'$ prior timestamps to predict $T$ following, in other words, given graph $\mathcal{G}$ and complementary information $\mathcal{I}$ (e.g., day of the week), we aim to find the best mapping function $f$ from space $\mathbb{R}^{N\times P\times T'}$ to $\mathbb{R}^{N\times Q\times T}$:
 \begin{multline}
    (\mathcal{G}; \mathcal{I}; \mathbf{X}^{(t-T'+1)}, \mathbf{X}^{(t-T')}, \cdots, \mathbf{X}^{(t)})\\ 
    \xrightarrow{f}(\hat{\mathbf{X}}^{(t+1)},\cdots, \hat{\mathbf{X}}^{(t+T-1)}, \cdots, \hat{\mathbf{X}}^{(t+T)})\notag
 \end{multline}
In this work, we focus on using historical one-hour speed data for 15min, 30min, and 60min ahead speed forecasting $(P=Q=1)$. More specifically, when the speed readings are aggregated into 5 minutes windows, it turns out to be: 
 \begin{multline}
    (\mathcal{G}; \mathcal{I}; \mathbf{X}^{(t-11)}, \mathbf{X}^{(t-10)}, \cdots, \mathbf{X}^{(t)})\\
    \xrightarrow{f}(\hat{\mathbf{X}}^{(t+3)}, \hat{\mathbf{X}}^{(t+6)}, \hat{\mathbf{X}}^{(t+12)}) \notag    
 \end{multline}
 
\subsection{Causal discovery}
\label{causalex}
We need to clarify that, though causal discovery comes from statistical ideology, it indeed surpasses the latter. One ordinary example is that, despite the strong positive correlation between ice cream sales on the beach and the number of drownings, taking ice cream does not lead to drowning. Factors behind this correlation are the season and the temperature. Therefore, causal structure discovery is necessary to reveal the faithful relation behind observational data.

% We conduct causal structure discovery experiments on the PEMS-BAY dataset to identify relation between stations. We hold the view that, in a speed prediction scenario, changes in the external environment and events such as accidents can be reflected in speed changes. 

The causal structure discovery analyzes the causal relation between different variables. In this work, we adopt ICD \cite{rohekar2021iterative} as our causal discovery program. The input of ICD is a set of variable values, and the output is the graph adjacency matrix, in which each element represents the importance of the causal relation between variables. The algorithm is only implemented on the PEMS-BAY dataset since there are too many missing values in METR-LA (8.11\%) than in PEMS-BAY (0.003\%). The dataset will be further described in Section \ref{datasetdesp}.

\begin{figure}[ht]
\centering
\includegraphics[width=\linewidth]{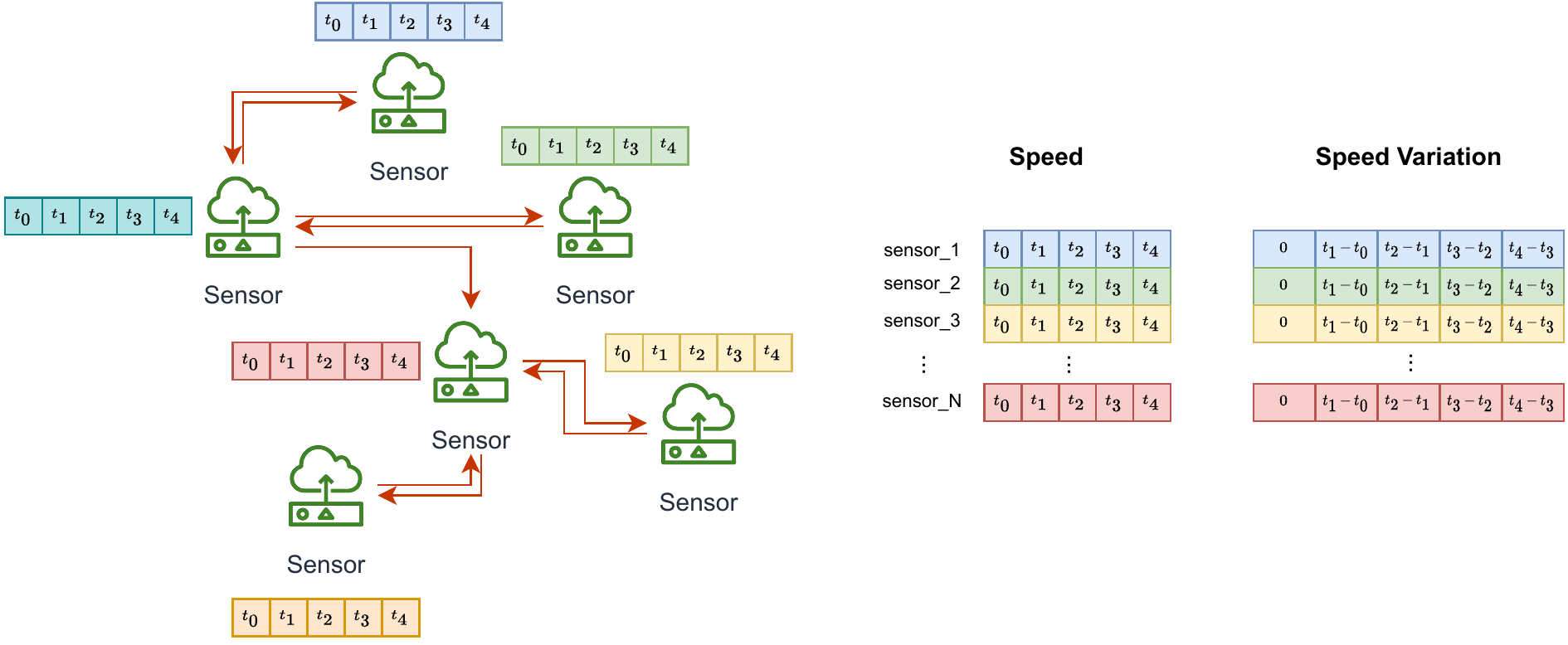}
%\framebox[4.0in]{$\;$}
% \fbox{\rule[-.5cm]{0cm}{4cm} \rule[-.5cm]{4cm}{0cm}}
\caption{Example of deriving speed variation of each node with 5 timestamps as input. }
\label{speedVar}
\end{figure}

Defining variables is a crucial step for causal discovery. Compared to the speed itself, the speed variation overall represents better the changes in traffic volume as well as the occurrence of external events, as it more clearly expresses the changes in traffic. Figure \ref{speedVar} represents road sensors as directed graph nodes. The speed detected by each sensor is a time series, and speed changes are obtained by taking the difference between adjacent time slices. What's more, Appendix \ref{distributiondata} shows that speed variation is a natural Gaussian distribution, which is necessarily required for ICD.

Given node $v_i$, we note $\mathbb{I}_1(i), \mathbb{O}_1(i)$ the set of its first-order in-neighbors/out-neighbors and $\mathbb{I}_2(i), \mathbb{O}_2(i)$ its second-order in-neighbors/out-neighbors. They can be formally formulated as below. 
\begin{gather}
    \mathbb{I}_1(i) = \{ j|(j, i)\in\mathcal{E}, i\neq j\}\\
    \mathbb{O}_1(i) = \{ j|(i, j)\in\mathcal{E}, i\neq j\}\\
    \mathbb{I}_2(i) = \{ j|\exists k, k\not\in \{i, j\}, (j, k)\in\mathcal{E}, (k, i)\in\mathcal{E},i\neq j\}\\
    \mathbb{O}_2(i) = \{ j|\exists k, k\not\in \{i, j\}, (i, k)\in\mathcal{E}, (k, j)\in\mathcal{E},i\neq j\}
\end{gather}
At the graph level, the input is denoted as $\mathbf{X}$, the integrated features along the adjacency matrix of first-order in-neighbors/out-neighbors are denoted as $\mathbf{I}_1, \mathbf{O}_1$ , which can be mathematically expressed as below.  
\begin{gather}
    \mathbf{W}_{I1} = \mathbf{W}^T-\mathbf{I}, \mathbf{W}_{O1} = \mathbf{W}-\mathbf{I}\\ 
     \begin{aligned}
    \mathbf{I}_1 = (\text{diag}(\mathbf{W}_{I1}\mathbf{1})^{-1}\mathbf{W}_{I1})\mathbf{X} 
    \end{aligned} \\
     \mathbf{O}_1 = (\text{diag}(\mathbf{W}_{O1}\mathbf{1})^{-1}\mathbf{W}_{O1})\mathbf{X} \label{curve} 
\end{gather}
For second order in-neighbors/out-neighbors, they are denoted as $\mathbf{I}_2, \mathbf{O}_2$: 
\begin{gather}
    \begin{aligned}
    \mathbf{W}_{I2} = \mathbf{W}_{I1}\mathbf{W}_{I1}, \text{diag}(\mathbf{W}_{I2})=0
    \end{aligned} \\
    \mathbf{W}_{O2} = \mathbf{W}_{O1}\mathbf{W}_{O1}, \text{diag}(\mathbf{W}_{O2})=0\\ 
    \begin{aligned}
    \mathbf{I}_2 = (\text{diag}(\mathbf{W}_{I2}\mathbf{1})^{-1}\mathbf{W}_{I2})\mathbf{X}
    \end{aligned} \\
    \mathbf{O}_2= (\text{diag}(\mathbf{W}_{O2}\mathbf{1})^{-1}\mathbf{W}_{O2})\mathbf{X}
\end{gather}

Given time $t$ and node $i$, for each timestamp $t_0$ between $t$ and $t+30\min$, we define variables to be the \textbf{speed variation} of $X^{(t_0)}_i, I_{1i}^{(t_0)}, O_{1i}^{(t_0)}, I_{2i}^{(t_0)}, O_{2i}^{(t_0)}$. Since the PEMS-BAY dataset aggregates traffic speed readings into 5 minutes windows, causal variables will be stored in a set of 6-time slices, resulting in 30 causal inferred variables below. The output of ICD is thus a $30\times 30$ matrix $\mathbf{C}$, where $C_{i,j}$ represents the causal relation between variable $i$ and variable $j$.
\begin{align}
\mathbb{V} = &\{V_0, V_1, \dots,V_{30} \}\notag \\
= & \{X^{(t)}_i, I_{1i}^{(t)}, O_{1i}^{(t)}, I_{2i}^{(t)}, O_{2i}^{(t)},\notag\\
&\cdots, X^{(t+5)}_i, I_{1i}^{(t+5)}, O_{1i}^{(t+5)}, I_{2i}^{(t+5)}, O_{2i}^{(t+5)}\}
\label{causVar}
\end{align}
We randomly sample data in the dataset and then process it to obtain the above 30 causal variables. To reduce accidental error, we sample 2000 samples each time as input of ICD, and repeat 100 times to sum the output matrix $\mathbf{C}$. 

A strong link between $v_i, \mathbb{I}_1(i)$ and $\mathbb{O}_1(i)$ is revealed from the causal discovery program, which is double-checked with statistical correlation and case analysis in Appendix \ref{causalrlt}.

\section{Methodology} \label{methods}
In this section, we start by formulating the studied problem, then present an overview of our model structure, and end with a detailed introduction of its crucial component layers.

\subsection{Model structure overview}
\label{modelsection}

\begin{figure}[ht]
\centering
\includegraphics[width=\linewidth]{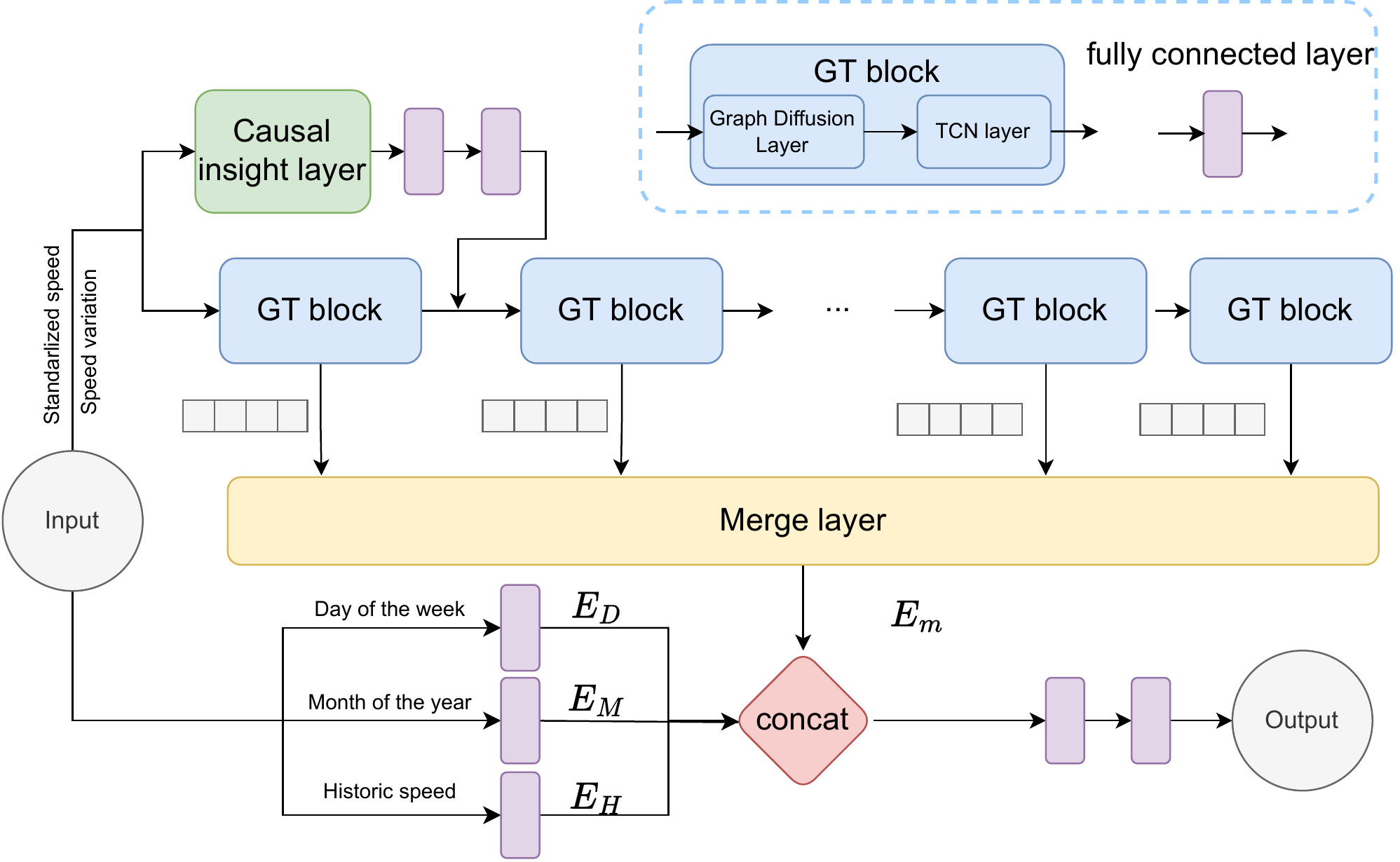}
%\framebox[4.0in]{$\;$}
% \fbox{\rule[-.5cm]{0cm}{4cm} \rule[-.5cm]{4cm}{0cm}}
\caption{System architecture for the \textit{Graph Spatial-Temporal Network Based on Causal Insight}. Time series are fed into the causal insight layer and a series of GT blocks. Skip connections are used at the end of each GT block to guard useful information. Periodic features are finally merged to make predictions. }
\label{model}
\end{figure}

Figure \ref{model} is a brief illustration of GT-CausIn structure. The main components of GT-CausIn model consist of the causal insight layer, graph diffusion layer, TCN layer, merge layer, and inherent feature layer. A detailed explanation of each layer can be found in Section \ref{modules}.

Graph diffusion layers facilitate information propagation through spatial nodes, but not inter-time series communication. TCN layer, which focuses on convolution along the temporal dimension, becomes complementary to graph diffusion layers. For the sake of expression simplicity, we name GT block as the combination of one graph diffusion layer and one TCN layer. Serializing multiple GT blocks can thus disclose inter-spatiotemporal correlations. 

The input speed, together with its differentiated variant, is fed into a causal insight layer, where the neighbor-level information is carefully investigated. Two fully connected layers are stacked to better fit vector space after the first GT block.

The number of GT block $L$ is not limited to a certain value,  bigger $L$ means a bigger receptive field, but model complexity is increased at the same time and important long-sequence information may be dispersed, the effects of $L$ will be further studied in Section \ref{ablasection}. Skip connections are added to stack hidden states learned at different scales, which eases feature propagation for short-term and long-term prediction.

\subsection{Component layers}   
\label{modules}
\textbf{Causal insight layer} 

\begin{figure}[ht]
\centering
\includegraphics[width=\linewidth]{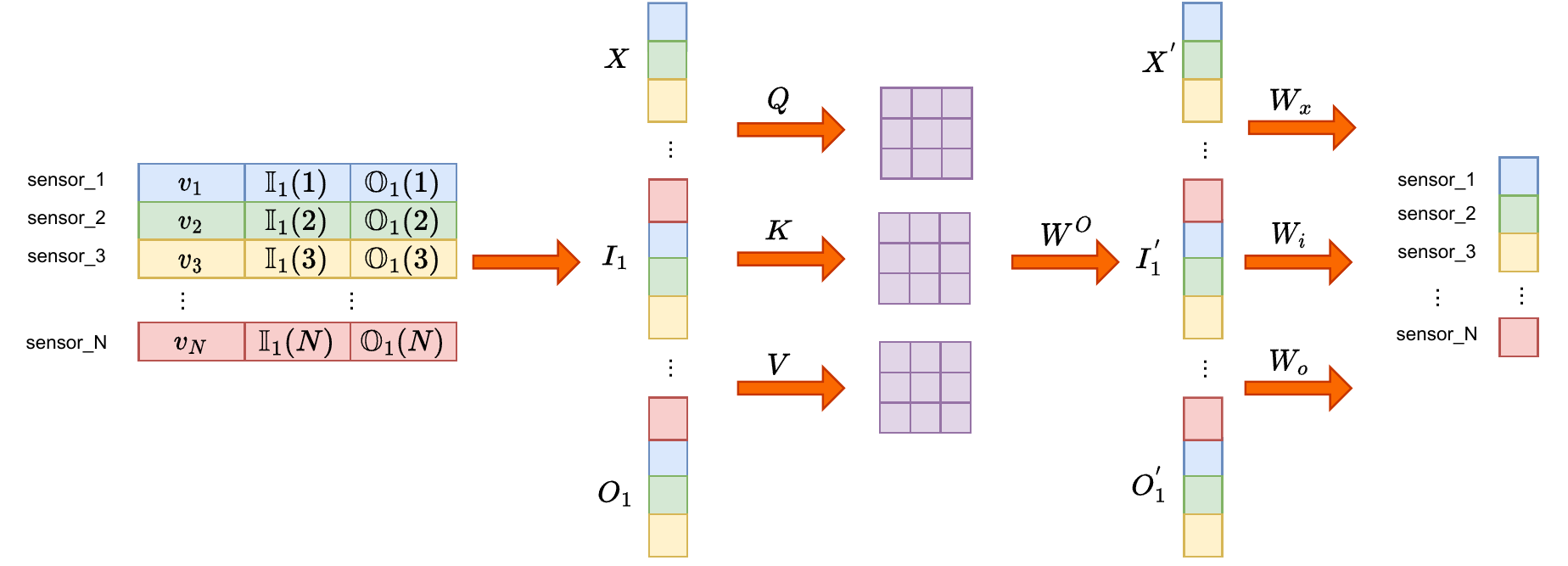}
%\framebox[4.0in]{$\;$}
% \fbox{\rule[-.5cm]{0cm}{4cm} \rule[-.5cm]{4cm}{0cm}}
\caption{A brief illustration of the causal insight layer. Node features are first concatenated with neighbor embedding features, then pass to an attention layer to reveal interaction influence. Three matrices are followed to learn individual influence on each node.}
\label{causLayer}
\end{figure}

From Section \ref{causalex}, we get the information that special attention on the first-order in-neighbors/out-neighbors need to be paid. The causal insight layer is therefore designed to embed this prior knowledge. To be cautious with the possible mutual influence between $\mathbb{I}_1(i)$, $\mathbb{O}_1(i)$, and different nodes $v_i$, we adopt a self-attention mechanism to learn inherent relations between them. Neighbor-level and node-level embedding spaces are followed to convert to individual node spaces. 

The speed variation is concatenated with speed to form the input of the causal insight layer. With the speed variation of each node and road connection matrix, we can deduce the embedding of neighbor levels. Therefore, the whole process is formulated as below and is depicted in Figure \ref{causLayer}.
\begin{align}
    \mathbf{I}_1 = \mathbf{D}_{I1}^{-1}&(\mathbf{W}^T-\mathbf{I})\mathbf{X}, \mathbf{O}_1 = \mathbf{D}_{O1}^{-1}(\mathbf{W}-\mathbf{I})\mathbf{X}\\ 
    \mathcal{S} &= \mathbf{X}\bigoplus\mathbf{I}_1\bigoplus\mathbf{O}_1 \label{ab0}\\
    \mathcal{S}^{'} &= \mathbf{X}^{'}\bigoplus\mathbf{I}_1^{'}\bigoplus\mathbf{O}_1^{'} \notag\\
    &= \text{attention}(\mathcal{S}\mathbf{W}_q, \mathcal{S}\mathbf{W}_k, \mathcal{S}\mathbf{W}_v)\mathbf{W}^O \label{ab1}\\
    \text{output}&=\mathbf{W}_x\mathbf{X}^{'}+\mathbf{W}_i\mathbf{I}_1^{'}+\mathbf{W}_o\mathbf{O}_1^{'} \label{ab2}
\end{align}
where $\mathbf{D}_{I1}=\text{diag}((\mathbf{W}^T-\mathbf{I})\mathbf{1}), \mathbf{D}_{O1}=\text{diag}((\mathbf{W}-\mathbf{I})\mathbf{1})$, $\mathbf{1}$ is the all one vector.

\textbf{Graph diffusion layer}  Spatial dependency is modeled in this part. Consider a directed graph $\mathcal{G}$, the diffusion process is modeled by a Markov process with the transition matrix $\mathbf{D}_{O}^{-1}\mathbf{W}$ and a starting probability $\alpha\in [0, 1]$ ($\mathbf{D}_{O}=\text{diag}(\mathbf{W}\mathbf{1})$). Teng \textit{et al.} \cite{teng2016scalable} show that the ultimate distribution is converged to a stationary contribution:
\begin{equation}
    \mathcal{P}=\sum_{k=0}^{\infty}\alpha(1-\alpha)^k(\mathbf{D}_{O}^{-1}\mathbf{W})^k
\end{equation}
where $k$ is the diffusion step. Based on this theory, Li \textit{et al.} \cite{li2017diffusion} present a formal expression of the graph signal diffusion process, which truncates the diffusion process to be $K$ steps and assigns trainable weights for each step.

For each input feature $p\in\{1,\cdots,P\}$ and output feature $q\in\{1,\cdots,Q\}$, 
\begin{align}
    &\mathbf{H}_{:,q}=\mathbf{X}_{:,p\star\mathcal{G}}f_{\mathrm{\theta}_{q,p,:,:}} 
    \notag\\
    &=\sum_{k=0}^{K-1}(\theta_{q,p,k,1}(\mathbf{D}_O^{-1}\mathbf{W})^k+\theta_{q,p,k,2}(\mathbf{D}_I^{-1}\mathbf{W}^T)^k)\mathbf{X}_{:,p} 
\end{align}

where $\mathbf{X}$ is the layer input, $\mathbf{H}$ the layer output, $\bm{\theta}\in\mathbb{R}^{Q\times P\times K\times 2}$ the filter parameter, $\mathbf{D}_I=\text{diag}(\mathbf{1}\mathbf{W})$ and $\mathbf{W}^T$ the transpose of graph adjacency matrix $\mathbf{W}$.

Different from the popular spectral GCN method proposed by Kipf \textit{et al.} \cite{kipf2016semi}, which is only applicable for undirected graphs, graph diffusion layers simulate well the diffusion process of directed graphs, and become a natural choice for traffic scenarios. 

\textbf{TCN layer} We adopt the dilated causal convolution described in Yu \textit{et al.} \cite{tcnyu} as our TCN layer. In contrast to RNN-based networks, dilated causal convolution can capture information over long-range sequences while getting rid of the gradient explosion problem. Concerning dilated causal convolution, stacking dilated layers ensures the exponential receptive field, and zero padding guarantees the temporal causal order since only historical information is involved to predict the current time step. Considering input feature dimension as $P$, output dimension as $Q$, given a time-series input $\mathbf{X}\in\mathbb{R}^{N\times P\times T}$ and a filter with trainable parameter $\mathrm{\theta}\in\mathbb{R}^{Q\times P\times K}$, the dilated causal convolution at time $t$ can be formulated as:
\begin{equation}
    \mathbf{H}_{:,q,t}=\mathbf{X}_{:,p, :}\star f_{\theta_{q,p,k}}(t)=\sum_{k=0}^{K-1}\theta_{q,p,k}x_{:, p, t-d*k}
\end{equation}
where $K$ is the kernel size, $d$ is the dilation factor, which decides the skipping distance. 

\textbf{Merge layer}    Although GT block extracts dependencies in the temporal and spatial dimension, information may still be lost for such a long series forecasting task. As a result, we keep the embedding after each GT block and concatenate them to enhance information flow over great lengths.

\begin{figure}[!t]
\subfloat[Station 400714]{
  \includegraphics[width=0.325\linewidth]{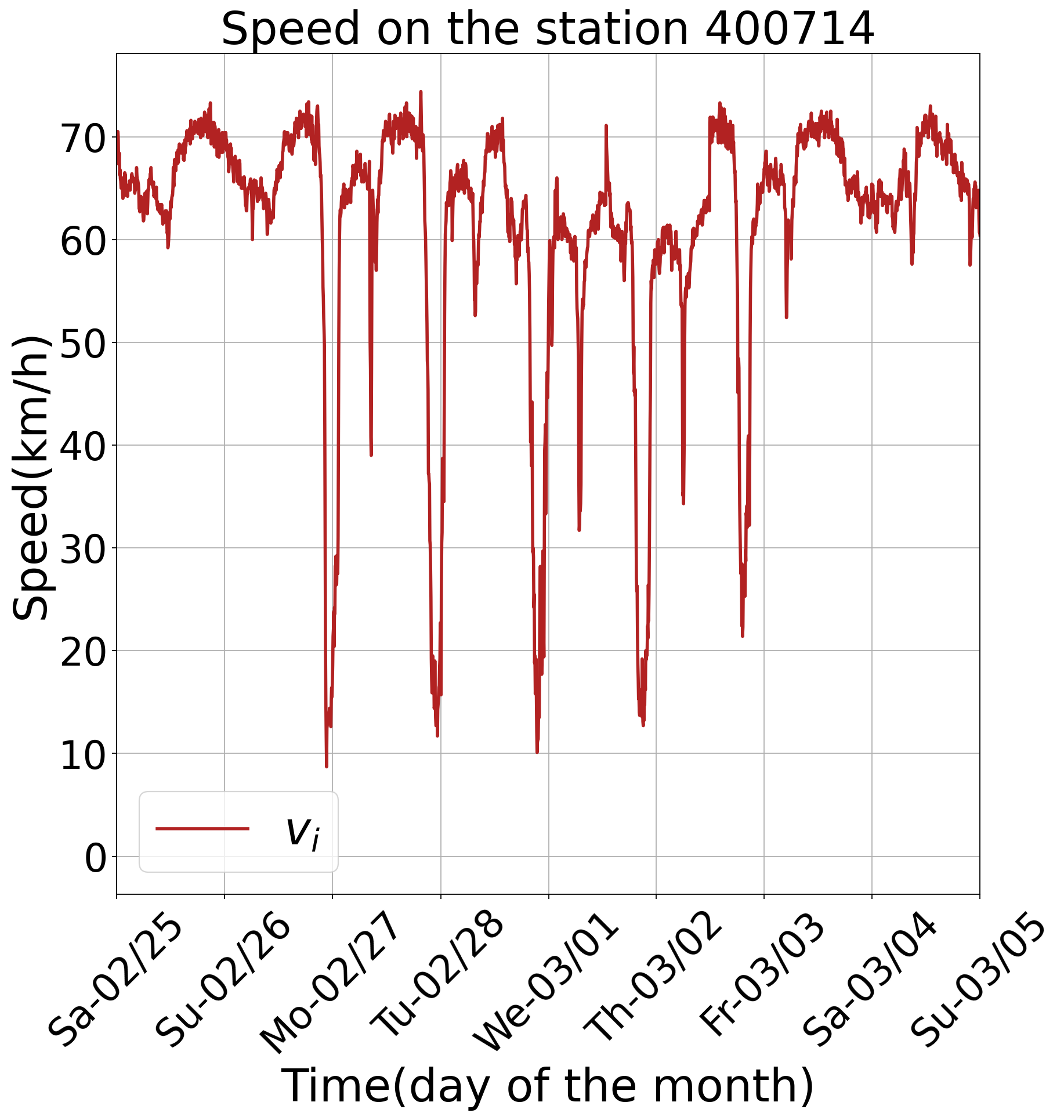}}
\subfloat[Station 400147]{
  \includegraphics[width=0.325\linewidth]{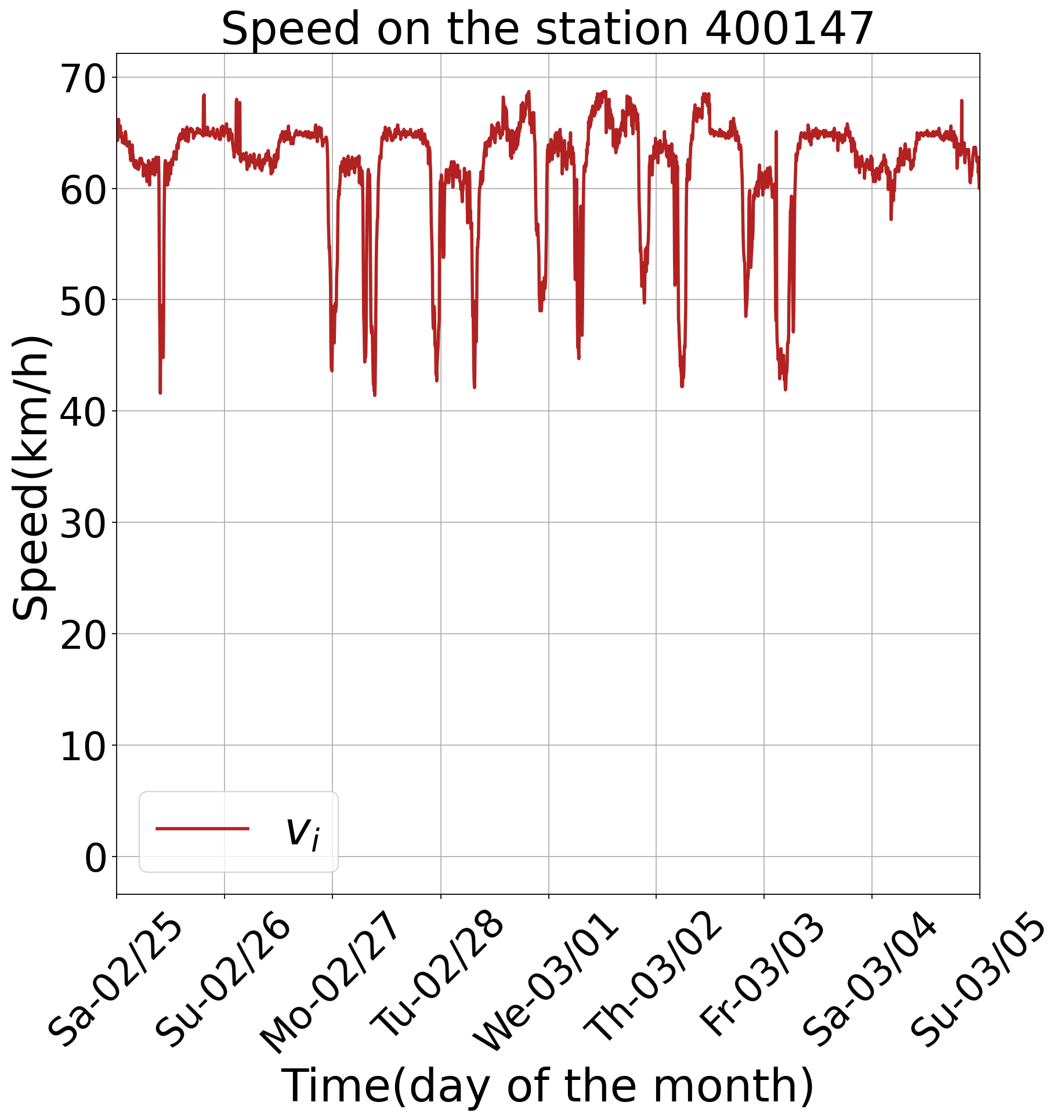}}
\subfloat[Station 401403]{
  \includegraphics[width=0.325\linewidth]{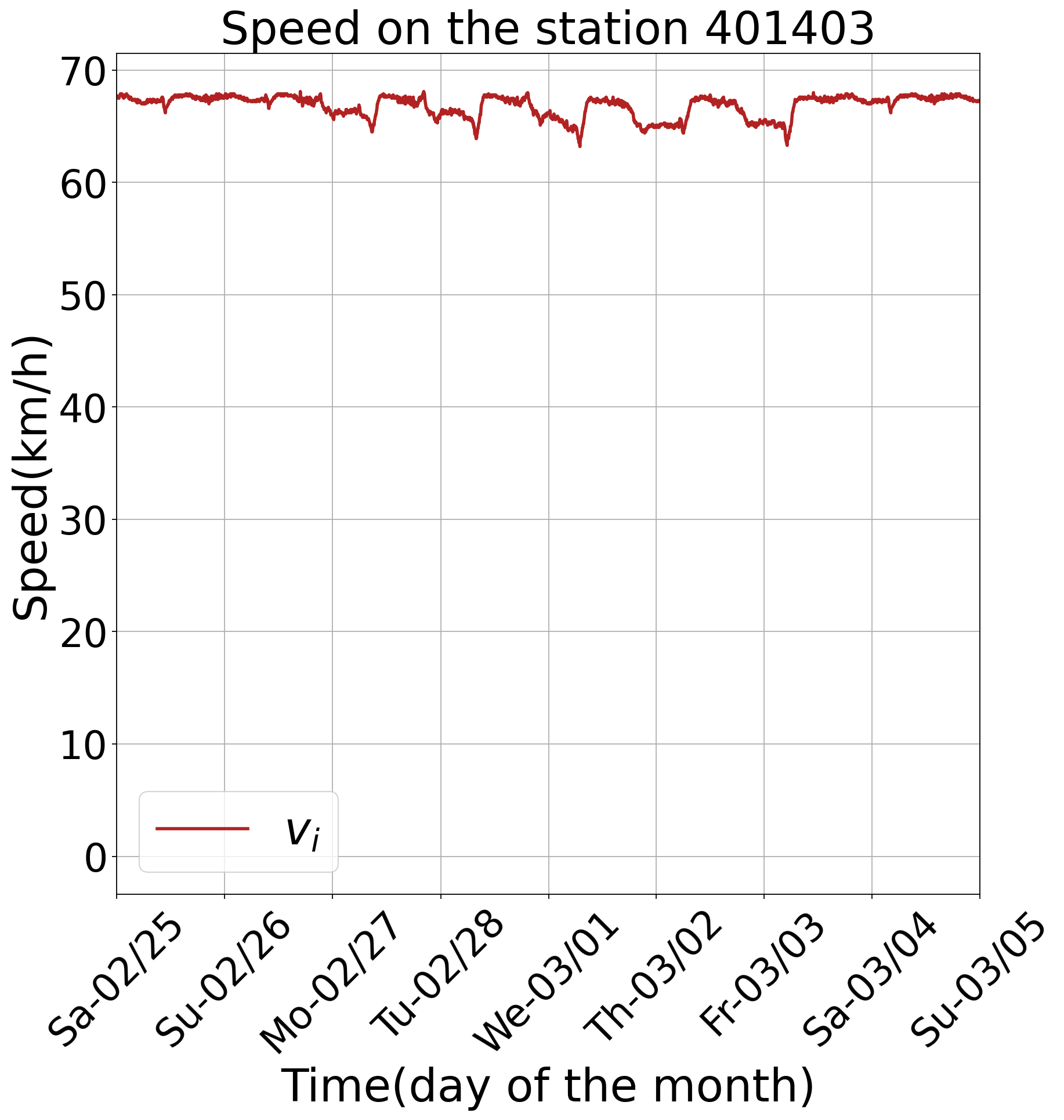}}
\caption{Periodic speed changes of stations of PEMS-BAY within one week.}
\label{speedI0}
\end{figure}

\textbf{Inherent feature layer} We notice that speed changes periodically for stations as shown in Figure \ref{speedI0}. Hence, we extract the day of the week, the month of the year, and historic speed as inherent features. Day of the week and month of the year features are represented as one-hot vectors while historic speed is the average speed of past days at the same moment. Since people's travel routines are very different on weekdays and weekends, we only use the past five weekday records or the past two weekend records to calculate historic speed.

The inherent features are integrated as follows: each feature goes through a dense layer to get embedding $\mathbf{E}_D$ (day of the week), $\mathbf{E}_M$ (month of the year), and $\mathbf{E}_H$ (historic speed). Together with embedding $\mathbf{E}_m$ obtained from merge layer, all features are fused into a long feature matrix by concatenation, e.g.,  $\mathbf{F}=\mathbf{E}_m\bigoplus\mathbf{E}_D\bigoplus\mathbf{E}_M\bigoplus\mathbf{E}_H$, then we stack two fully connected layers upon $\mathbf{F}$ to ensure that the output is of the desired shape.

\section{Experiments}

\subsection{Dataset description}\label{datasetdesp}
We conduct dense experiments on two public datasets: PEMS-BAY \cite{li2017diffusion} and METR-LA \cite{metr}. PEMS-BAY is collected by California Transportation Agencies (CalTrans) Performance Measurement System (PeMS). It contains 325 highway sensors of the Bay Area from Jan 1st 2017 to Jun 30th 2017. For METR-LA, 207 loop detectors on the highway of Los Angeles County are selected from Mar 1st 2012 to Jun 30th 2012. For both datasets, traffic speed readings are aggregated into 5 minutes windows. Based on the road connection map, we adopt the same way as Li \textit{et al.} \cite{li2017diffusion} to get the graph adjacency matrix. With pairwise road network distances between sensors, the adjacency matrix representing connection from any sensor $i$ to any sensor $j$ is expressed as:
\begin{equation}
    W_{i,j} = \exp{(\frac{-d(v_i, v_j)^2}{\sigma^2})} \text{if } d(v_i, v_j)\leq \kappa, \text{otherwise }0
\end{equation}
where $\sigma$ is the standard deviation, $\kappa$ the threshold, and $d(v_i,v_j)$ the distance between station $v_i$ and station $v_j$. For both datasets, 80\% of the data serves as the training set, 10\% as the test set, and the remaining 10\% as the validation set.

\subsection{Experiment setting}
All models are implemented with PyTorch \cite{pytorch} and trained with an Adam optimizer with an annealing learning rate. Experiments on dataset METR-LA are carried out on two Nvidia Tesla V100 servers and PEMS-BAY on two Nvidia Tesla A100 servers. The detailed parameter setting for all experiments executed is available in Appendix \ref{exsetting}.

All methods are evaluated based on three commonly used metrics in traffic forecasting, which are Mean Absolute Error (MAE), Mean Absolute Percentage Error (MAPE), and Root Mean Squared Error (RMSE). The formula of all metrics can be found in Appendix \ref{metrics}. MAE is used as the loss function. Missing values are linearly filled for training, and are excluded for results evaluation.

\subsection{Comparison with baselines}
\begin{table*}[!t]
\caption{Performance comparison with baseline models for traffic speed forecasting.}
\label{sota}
\begin{center}
\resizebox{\linewidth}{52mm}{
\begin{tabular}{lllllllllll}
\hline
\multirow{2}{*}{\bf Data} &\multirow{2}{*}{\bf Method} &\multicolumn{3}{c}{\bf 15min} &\multicolumn{3}{c}{\bf 30min} &\multicolumn{3}{c}{\bf 60min} \\&  &\multicolumn{1}{c}{MAE} &\multicolumn{1}{c}{RMSE} &\multicolumn{1}{c}{MAPE} &\multicolumn{1}{c}{MAE} &\multicolumn{1}{c}{RMSE} &\multicolumn{1}{c}{MAPE} &\multicolumn{1}{c}{MAE} &\multicolumn{1}{c}{RMSE} &\multicolumn{1}{c}{MAPE}  \\
\hline
\multirow{11}{*}{\rotatebox[origin=c]{90}{METR-LA}} 
&DCRNN &2.77 &5.38 &7.30\% &3.15 &6.45 &8.8\% &3.60 &7.59 &10.5\% \\
&Graph Wavenet &2.69 &5.15 &6.90\% &3.07 &6.22 &8.37\% &3.53 &7.37 &10.01\%\\
&GMAN &4.04 &8.53 &10.2\% &4.59 &9.85 &11.69\% &5.33 &11.21 &13.60\%\\
&ST-GRAT &2.60 &5.07 &6.61\% &3.01 &6.21 &8.15\% &3.49 &7.42 &10.01\% \\
&SLCNN &\bf 2.53 &5.18	&6.70\% & 2.88 &6.15 & 8.00\%	&3.30	&7.20	&9.70\%\\
&DGCRN &2.62 &\bf 5.01 &6.63\% &2.99 &6.05 &8.02\% &3.44 &7.19 &9.73\% \\
&DMSTGCN &2.85 &5.54 &7.54\% &3.26 &6.56 &9.19\% &3.72 &7.55 &10.96\% \\
&PGCN &2.70 &5.16 &6.98\% &3.08 &6.22 &8.38\% &3.54 &7.36 &9.94\% \\
\cline{2-11} &GT-NoCausIn &2.82 &5.52 &7.18\% &3.27 &6.71 &8.90\% &3.95 &8.24 &11.67\%\\
&GT-CausIn &2.61 &5.07 &\bf 6.60\% &\bf 2.73 &\bf 5.34 &\bf 7.09\% &\bf 3.06 &\bf 6.11 &\bf 8.30\%\\
&Improvement &-3.2\% &-1.2\% &0.2\% &5.2\% &11.7\% &11.4\% &7.3\% &15.0\% &14.4\%

\\\hline
\multirow{11}{*}{\rotatebox[origin=c]{90}{PEMS-BAY}} 
&DCRNN &1.38 &2.95 &2.90\% &1.74 &3.97 &3.90\% &2.07 &4.74 &4.90\% \\
&Graph Wavenet &1.30 &2.74 &2.73\% &1.63 &3.70 &3.67\% &1.95 &4.52 &4.63\%\\
&GMAN &1.34 &2.82 &2.81\% &1.62 &3.72 &3.63\% &1.86 & 4.32 &4.31\%\\
&ST-GRAT &1.29 &2.71 &2.67\% &1.61 &3.69 &3.63\% &1.95 &4.54 &4.64\% \\
&SLCNN &1.44 &2.90	&3.00\% &1.72	&3.81 &3.90\% &2.03 &4.53 &4.80\% \\
&DGCRN  &\bf 1.28 &2.69 &\bf 2.66\% &1.59 &3.63 &3.55\% &1.89 &4.42 &4.43\%\\
&DMSTGCN &1.33 &2.83 &2.80\% &1.67 &3.79 &3.81\% &1.99 &4.54 &4.78\%\\
&PGCN &1.30 &2.73 &2.72\% &1.62 &3.67 &3.63\% &1.92 &4.45 &4.55\%\\
\cline{2-11} &GT-NoCausIn &1.43 &2.85 &3.00\% &1.79 &3.84 &4.01\% &2.25 &4.90 &5.44\%\\
&GT-CausIn &1.30 &\bf 2.54 &2.68\% &\bf 1.47 &\bf 2.92 &\bf 3.10\% &\bf 1.70 &\bf 3.37 &\bf 3.68\%\\
&Improvement &-1.6\% &5.6\% &-0.8\% &7.5\% &19.6\% &12.7\% &8.6\% &22.0\% &14.6\%
\\\hline
\end{tabular}}
\end{center}
\end{table*}

We compare GT-CausIn with various popular traffic forecasting models in recent years, including (1) DCRNN: Diffusion Convolutional Recurrent Neural Network \cite{li2017diffusion}; (2) Graph Wavenet \cite{wavenet}, which develops an adaptive dependency matrix based on node embedding; (3)GMAN: Graph Multi-Attention Network for Traffic Prediction \cite{zheng2020gman}, of which the main structure is an encoder and a decoder both with several attention blocks; (4) ST-GRAT: Spatio-temporal Graph Attention Networks for Accurately Forecasting Dynamically Changing Road Speed \cite{stgrat}, which includes spatial attention, temporal attention, and spatial sentinel vectors; (5) SLCNN: Structure Learning Convolution Neural Network \cite{zhang2020spatio}, which learns the graph structure information with convolutional methods; (6) DGCRN: Dynamic Graph Convolutional Recurrent Network (DGCRN) \cite{li2021dynamic}, which filters node embedding to generate dynamic graph at each time step; (7) DMSTGCN: Dynamic and Multi-faceted Spatio-temporal Deep Learning for Traffic Speed Forecasting \cite{dmstgcn}, which provides a multi-faceted fusion module to incorporate the hidden states learned at different stages; (8) PGCN: Progressive Graph Convolutional Networks for Spatial-temporal Traffic Forecasting \cite{shin2022pgcn}, which constructs progressive adjacency matrices by learning in training and test phases.

Our approach is different from all the approaches above, we first use a causal discovery program to discover a general rule between node neighbors of different orders, then build a model named GT-CausIn to enhance prediction performance. Although the causality relation is discovered with dataset PEMS-BAY, it generalizes as well on the other dataset METR-LA. Besides, the proposed causal insight layer is integrated with directed graph diffusion layers to model the spatial dependency. TCN layers and skip connections are further used to discover the temporal dependency.
 
A comparison of GT-CausIn and other baseline models is shown in Table \ref{sota}. The causal insight layer is taken out from GT-CausIn and the rest model is named as GT-NoCausIn. All the baseline results are taken from its original paper if it is available and from \cite{shin2022pgcn} if not\footnote{Only GMAN and DMSTGCN authors implement experiments on different datasets.}.  

From Table \ref{sota}, we can observe a large margin between GT-NoCausIn and GT-CausIn , which shows that causal insights can significantly improve prediction accuracy. We also notice that models with dynamic training graph generally outperforms other models, however, GT-CausIn achieves the best overall result, especially for mid-term (30min) and long-term (60min) prediction. We need to remark that DCRNN \cite{li2017diffusion} shares similar graph diffusion layers and GMAN \cite{zheng2020gman} widely adopt the attention mechanism, yet our model exceeds them on all metrics, which justifies our model structure. 
 
\subsection{Ablation study}
\label{ablasection}
\textbf{Causal effects}    In GT-CausIn, we put special attention to relations between node $v_i$, $\mathbb{I}_1(i)$ and $\mathbb{O}_1(i)$, nonetheless, this conclusion drawn from the causal discovery program may be questioned. Therefore, we replace Equation (\ref{ab0}, \ref{ab1}, \ref{ab2}) as below to check its effectiveness, naming this model GT-BadCausIn.
\begin{align}
    \mathcal{S} &= \mathbf{X}\bigoplus\mathbf{X}\bigoplus\mathbf{X} \label{ablation}\\
    \mathcal{S}^{'} &= \mathbf{X}^{'}_{0}\bigoplus\mathbf{X}_1^{'}\bigoplus\mathbf{X}_2^{'} \notag\\
    &=\text{attention}(\mathcal{S}\mathbf{W}_q, \mathcal{S}\mathbf{W}_k, \mathcal{S}\mathbf{W}_v)\mathbf{W}^O \label{abbad}\\
    \text{output}&=\mathbf{W}_{x_0}\mathbf{X}_0^{'}+\mathbf{W}_{x_1}\mathbf{X}_1^{'}+\mathbf{W}_{x_2}\mathbf{X}_2^{'}
\end{align}
Therefore, GT-BadCausIn integrates no neighbor embedding in causal insight layer. 
\begin{table*}[!t]
\caption{Ablation study on causal effects.}
\label{abstudy0}
\begin{center}
\resizebox{\linewidth}{18mm}{
\begin{tabular}{lllllllllll}
\hline
\multirow{2}{*}{\bf Data} &\multirow{2}{*}{\bf Method} &\multicolumn{3}{c}{\bf 15min} &\multicolumn{3}{c}{\bf 30min} &\multicolumn{3}{c}{\bf 60min} \\&  &\multicolumn{1}{c}{MAE} &\multicolumn{1}{c}{RMSE} &\multicolumn{1}{c}{MAPE} &\multicolumn{1}{c}{MAE} &\multicolumn{1}{c}{RMSE} &\multicolumn{1}{c}{MAPE} &\multicolumn{1}{c}{MAE} &\multicolumn{1}{c}{RMSE} &\multicolumn{1}{c}{MAPE}  \\
\hline
\multirow{3}{*}{METR-LA}
&GT-BadCausIn &2.65 &5.17 &6.72\% &2.86 &5.71 &7.58\% &3.18 &6.50 &8.85\%\\
&GT-CausIn &2.61 &5.07 &6.60\% &2.73 &5.34 &7.09\% &3.06 &6.11 &8.30\%\\
&Improvement &1.5\% &1.9\% &1.8\% &4.5\% &6.5\% &6.5\% &3.8\% &6.0\% &6.2\%
\\\hline
\multirow{3}{*}{PEMS-BAY}
&GT-BadCausIn  &1.35 &2.63 &2.80\% &1.56 &3.13 &3.34\% &1.81 &3.64 &4.01\%\\
&GT-CausIn  &1.30 &2.54 &2.68\% &1.47 &2.92 &3.10\% &1.70 &3.37 &3.68\%\\
&Improvement &3.7\% &3.4\% &4.3\% &5.8\% &6.7\% &7.2\% &5.0\% &7.4\% &8.2\%
\\\hline
\end{tabular}}
\end{center}
\end{table*}

Experiments results are summarised in Table \ref{abstudy0}. The impact brought by neighbor embedding in the causal insight layer is positive for all metrics on both datasets, especially for mid-term (30min) and long-term (60min) predictions. The reason behind may be the propagation time of causal effects. Besides, we can observe that the improvement is smaller on METR-LA than on PEMS-BAY, which may result from: (1) a larger percentage of missing values in METR-LA (8.11\%) than in PEMS-BAY (0.003\%); (2) causal discovery program is only executed on the PEMS-BAY dataset.

\begin{figure}[ht]
\subfloat[MAE/15min]{
  \includegraphics[width=0.325\linewidth]{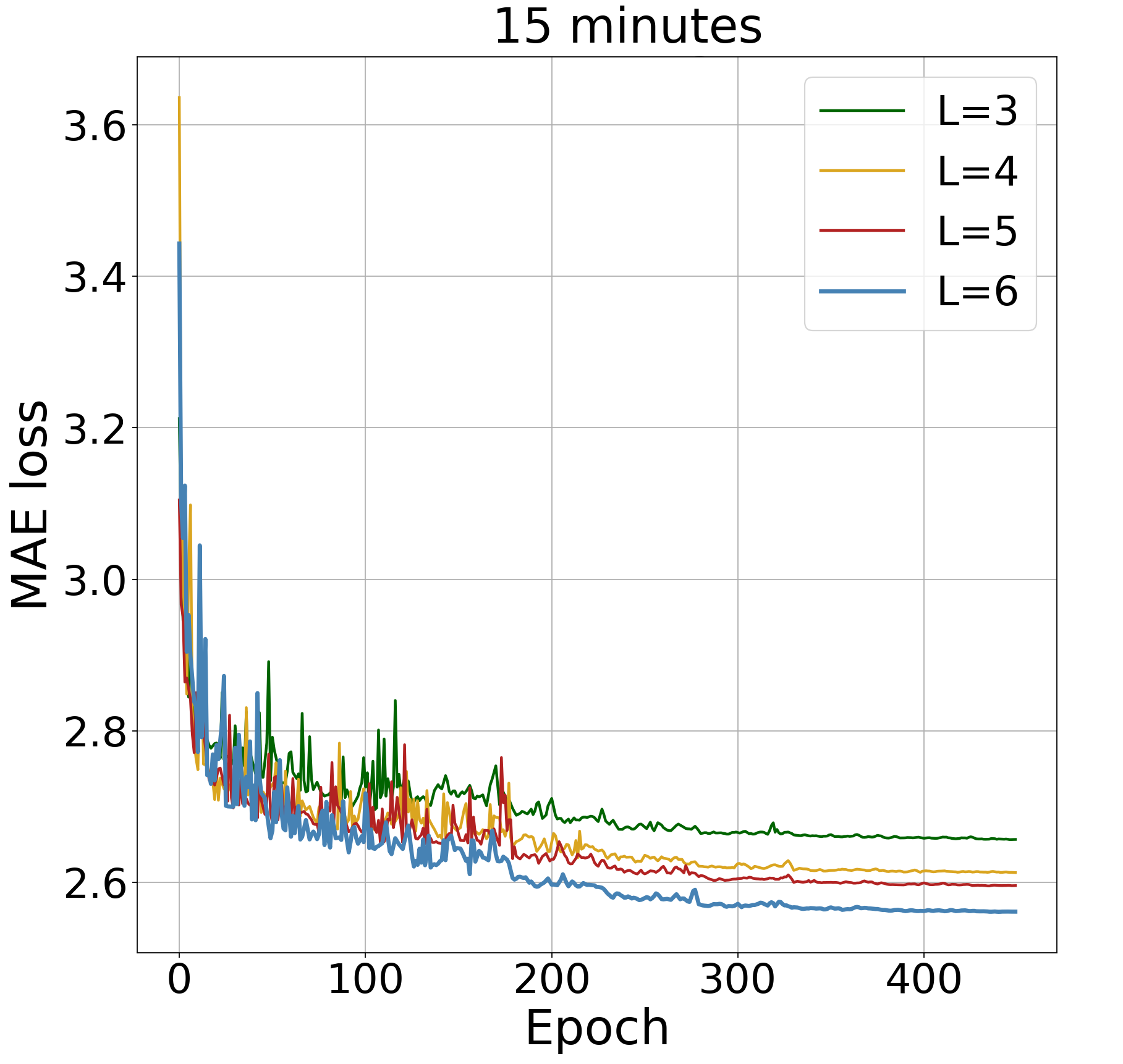}}
\subfloat[MAE/30min]{
  \includegraphics[width=0.325\linewidth]{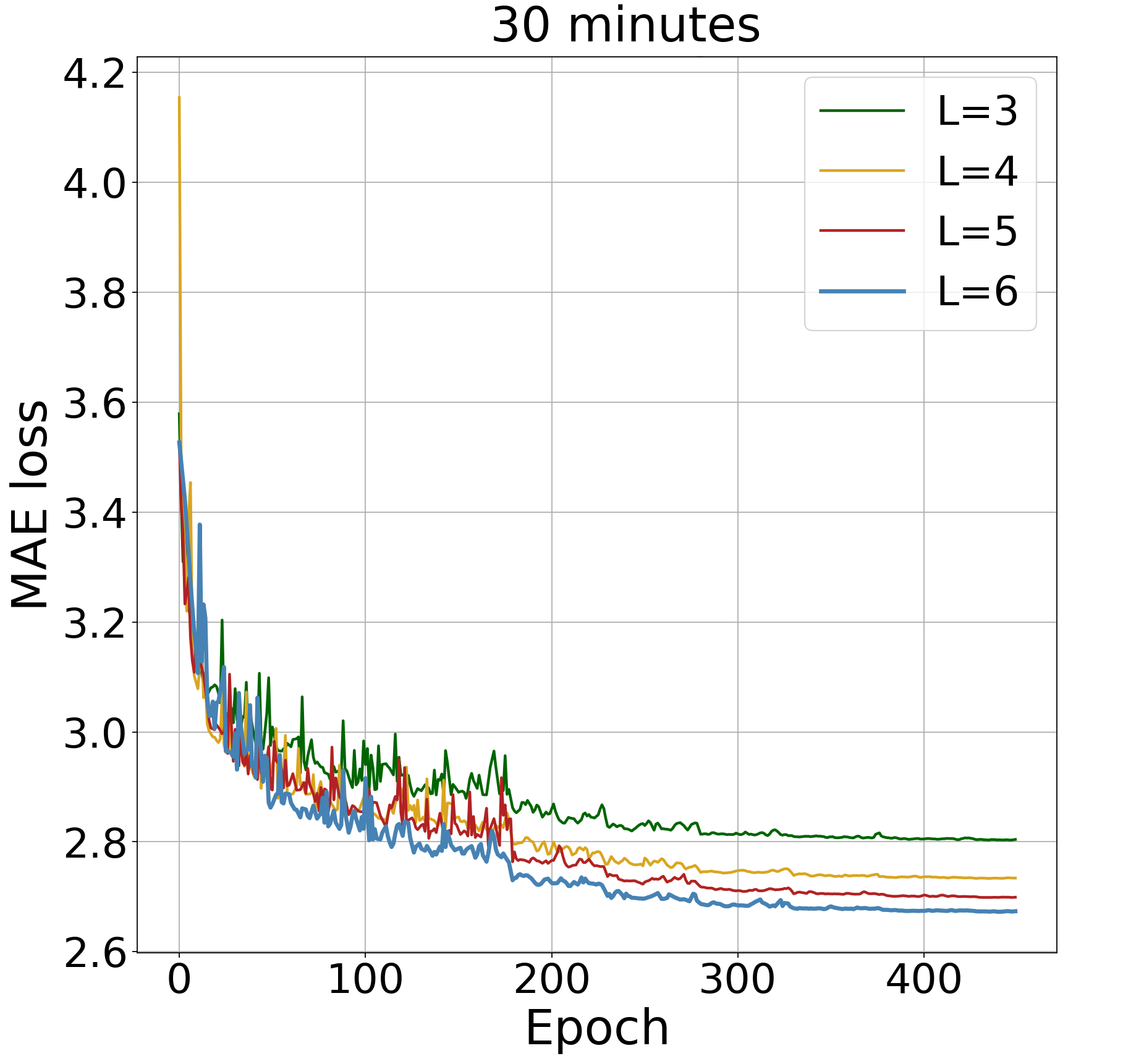}}
\subfloat[MAE/60min]{
  \includegraphics[width=0.325\linewidth]{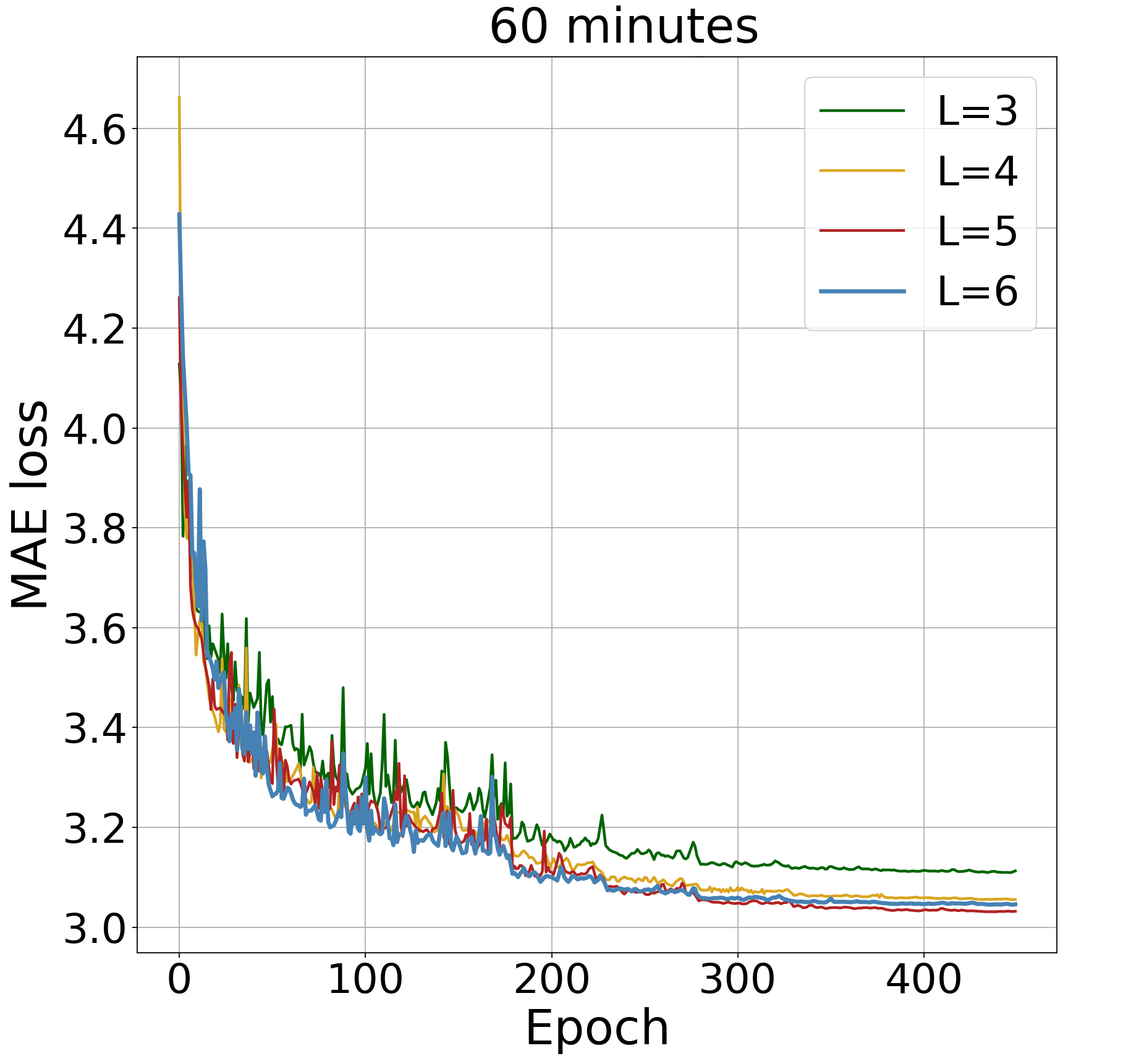}}
\caption{Parameter analysis of $L$ (number of stacking GT layers) on METR-LA dataset}
\label{ablationL}
\end{figure}

\textbf{Number of stacking GT blocks} In Section \ref{modelsection}, we introduced that the number of GT blocks $L$ is not limited to a certain value. Larger $L$ roughly corresponds to a larger spatiotemporal receptive field and enables the model to capture broader spatiotemporal dependency. We compare model performance with different $L \in\{3, 4, 5, 6\}$ on MAE of different time horizons in Figure \ref{ablationL}. We can observe a steady rise when $L$ goes up. Since the gap between $L=3$ and $L=4$ is more important than between $L=4$ and $L=5$, taking the increased model complexity into account, we adopt $L=4$ as our best parameter.

\subsection{Model  interpretation}
\label{modelInterpretation}
\begin{figure}[ht]
\subfloat[Station 71447]{
  \includegraphics[width=0.325\linewidth]{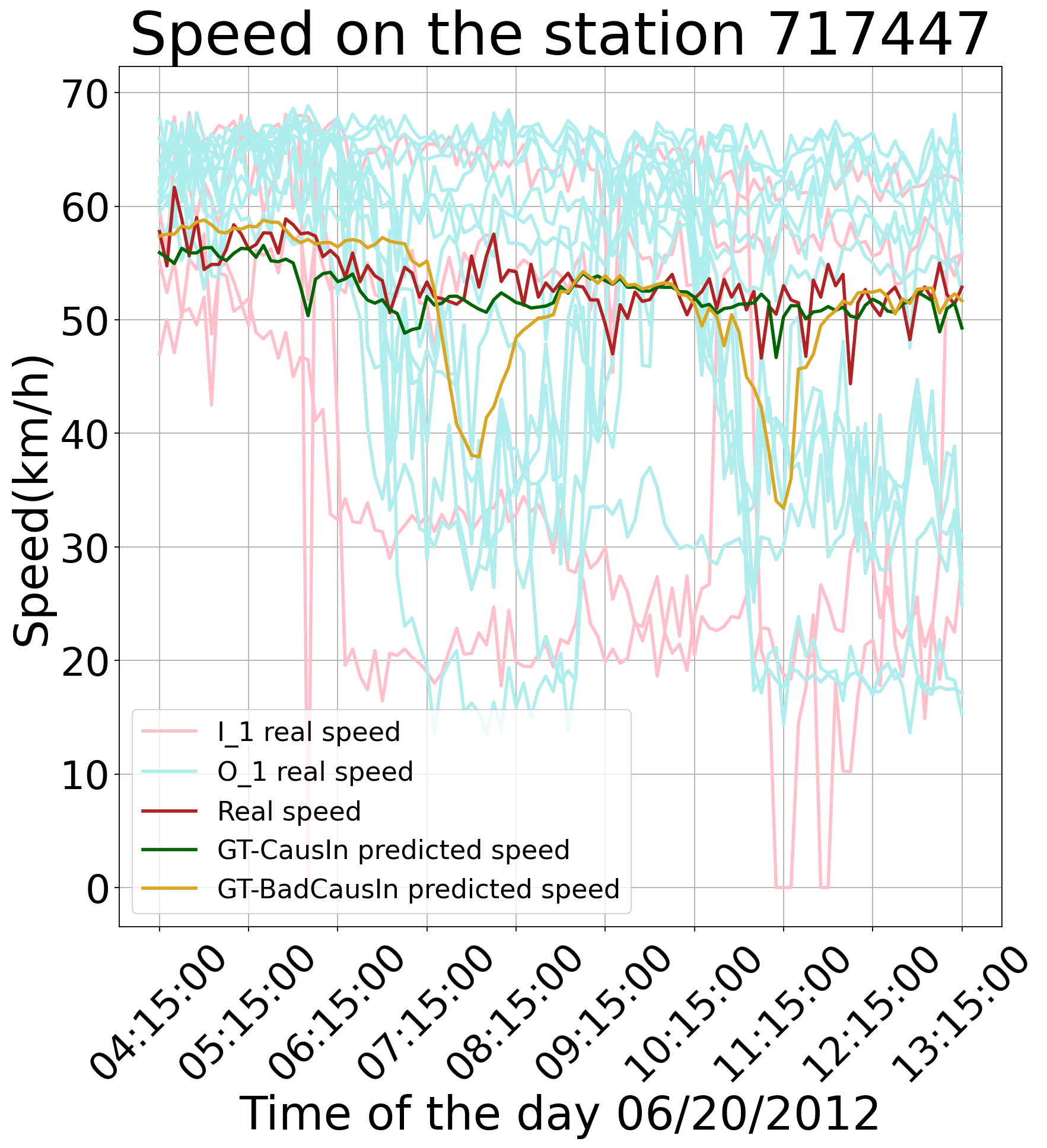}\label{subfiga}}
\subfloat[Station 773062]{
  \includegraphics[width=0.325\linewidth]{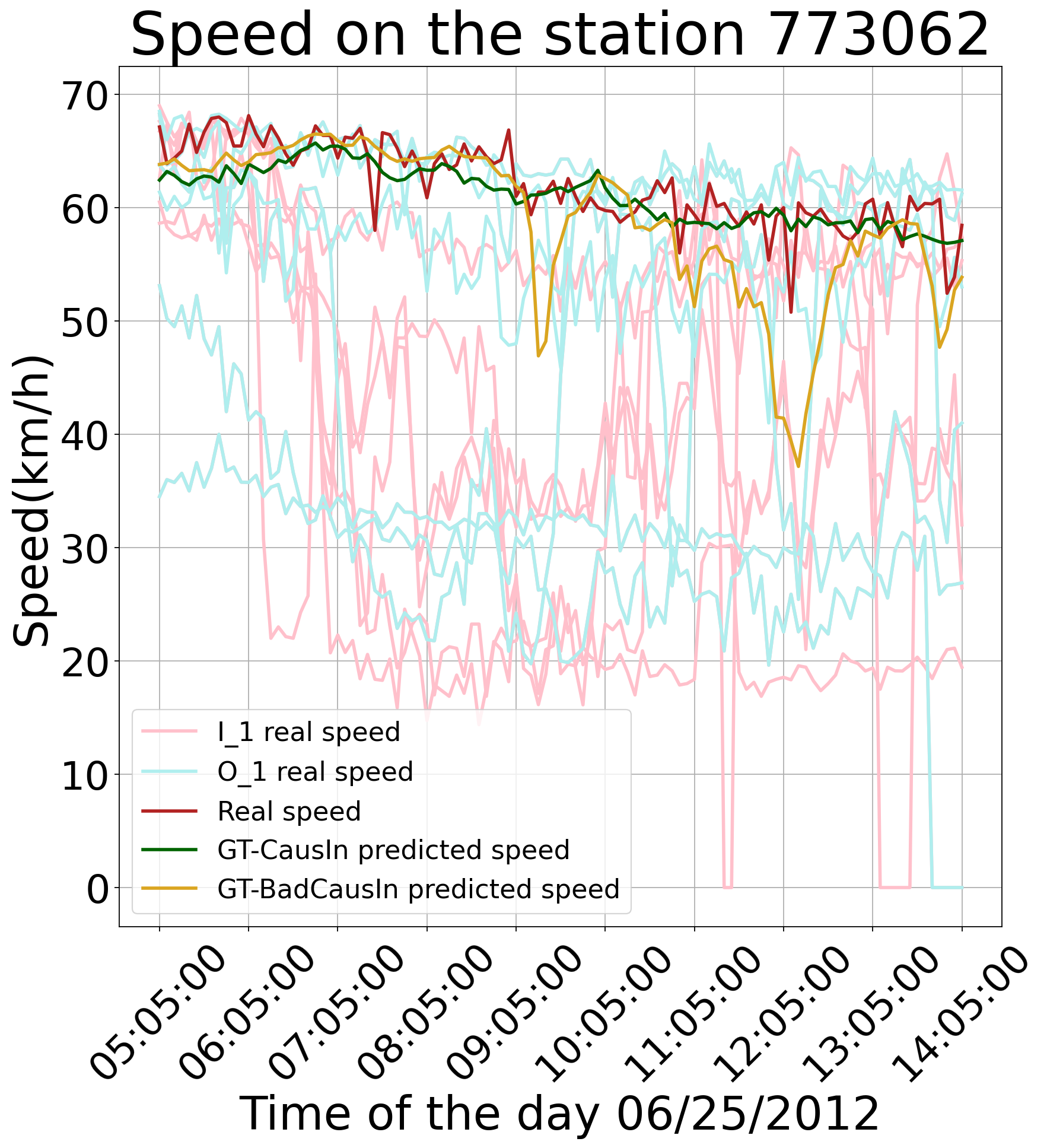}}
\subfloat[Station 716337]{
  \includegraphics[width=0.325\linewidth]{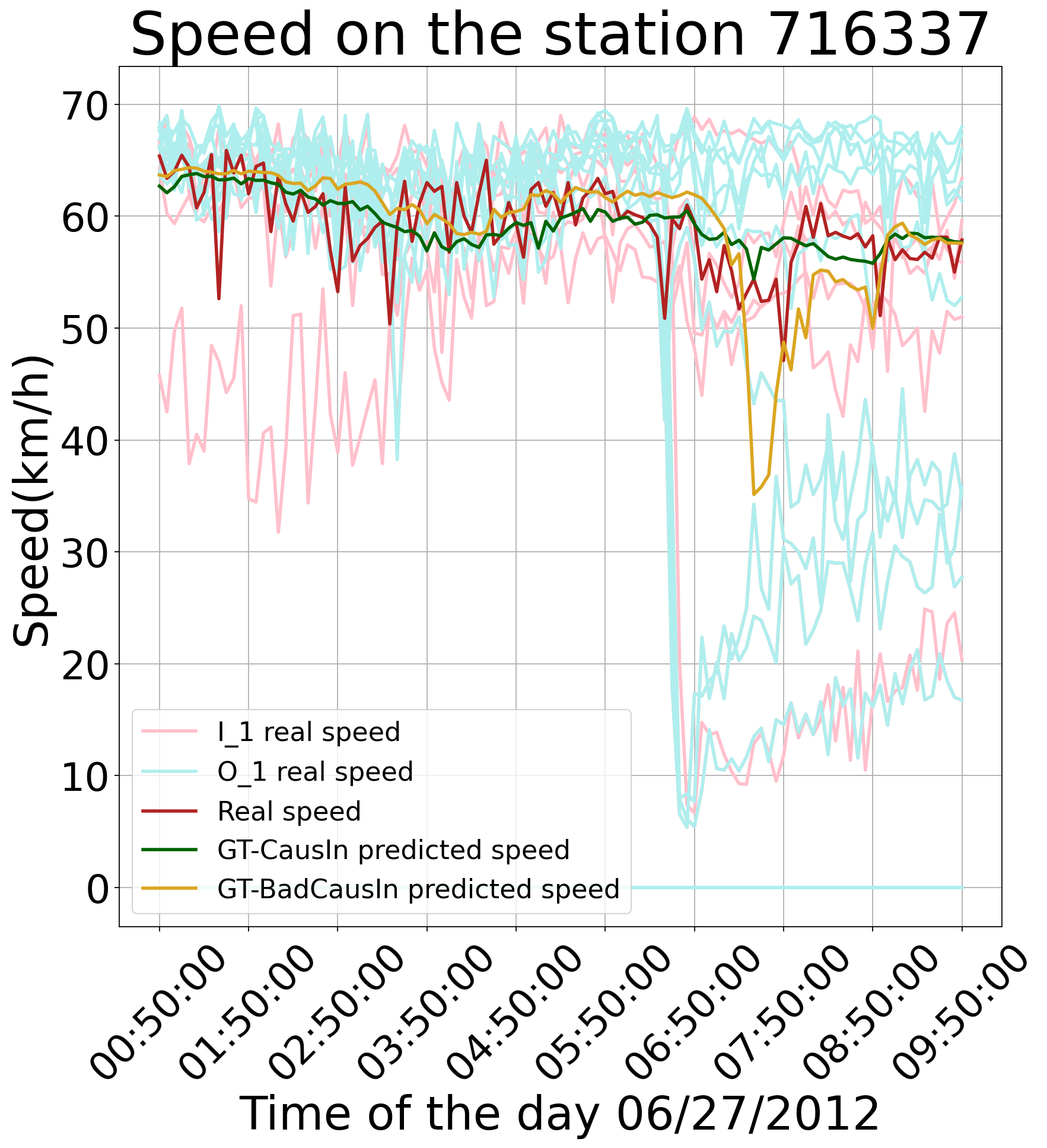}}
\caption{Figures (a), (b), and (c) offer a comparison between GT-CausIn and GT-BadCausin for prediction task on METR-LA, differing on the target station and target time. Each figure contains the ground truth of the considered station and its neighbors, as well as its 60min ahead prediction. It is notable that the predictions of GT-BadCausIn are heavily dependent on some neighbors, and are therefore swayed off the ground truth. The same cannot be said for GT-causIn, instead of being influenced by some stations, it has a broader outlook of its neighbors, resulting in a much more faithful and smooth prediction.}
\label{interprepic}
\end{figure}

To better understand our model, we compare it with GT-BadCausIn on 60min ahead prediction task. We shift the input window to get continuous predictions in Figure \ref{interprepic}, observing a notable performance margin between GT-CausIn and GT-BadCausIn. As described in Section \ref{ablasection}, the only structural difference between these two models is the dissimilar perspectives integrated. For GT-BadCausIn, only node-level information $\mathbf{X}$ is investigated. For GT-CausIn, neighbor-level information $\mathbf{I}$ and $\mathbf{O}$ are also included. This structural change results in different token attention scores of the causal insight layer and is responsible for this disparity, this is further investigated in Appendix \ref{attentionScore}.

\section{Conclusion}
In this paper, we implemented knowledge learned from a causal discovery program with deep learning models and proposed \textit{Graph Spatial-Temporal Network Based on Causal Insight} to capture spatiotemporal dependencies. Specifically, we serialize graph diffusion layers and TCN layers to capture dependencies between spatial flow and temporal flow, we also use skip connections to guarantee information propagation for long sequences. We creatively design a causal insight layer that focuses on stations, their first-order in-neighbors, and their first-order out-neighbors. When evaluating our model on two real-world traffic datasets, we achieved significantly better overall prediction than baselines. We also conducted ablation studies to show the effectiveness of causal knowledge and the influence of the stacking layer number. In the future, we plan to (1) apply the proposed model to other spatial-temporal forecasting tasks; (2) investigate scalable methods to apply to large-scale datasets.

% \bibliographystyle{IEEEtran}
% \bibliography{IEEE}

\appendices

\section{Causal result validation}
\label{causalrlt}
In this section, we carefully validate causal results obtained in Appendix \ref{causalex} in two ways: (1) coherence with the correlation between variables; (2) case analysis.

\begin{figure}[ht]
\centering
\includegraphics[width=\linewidth]{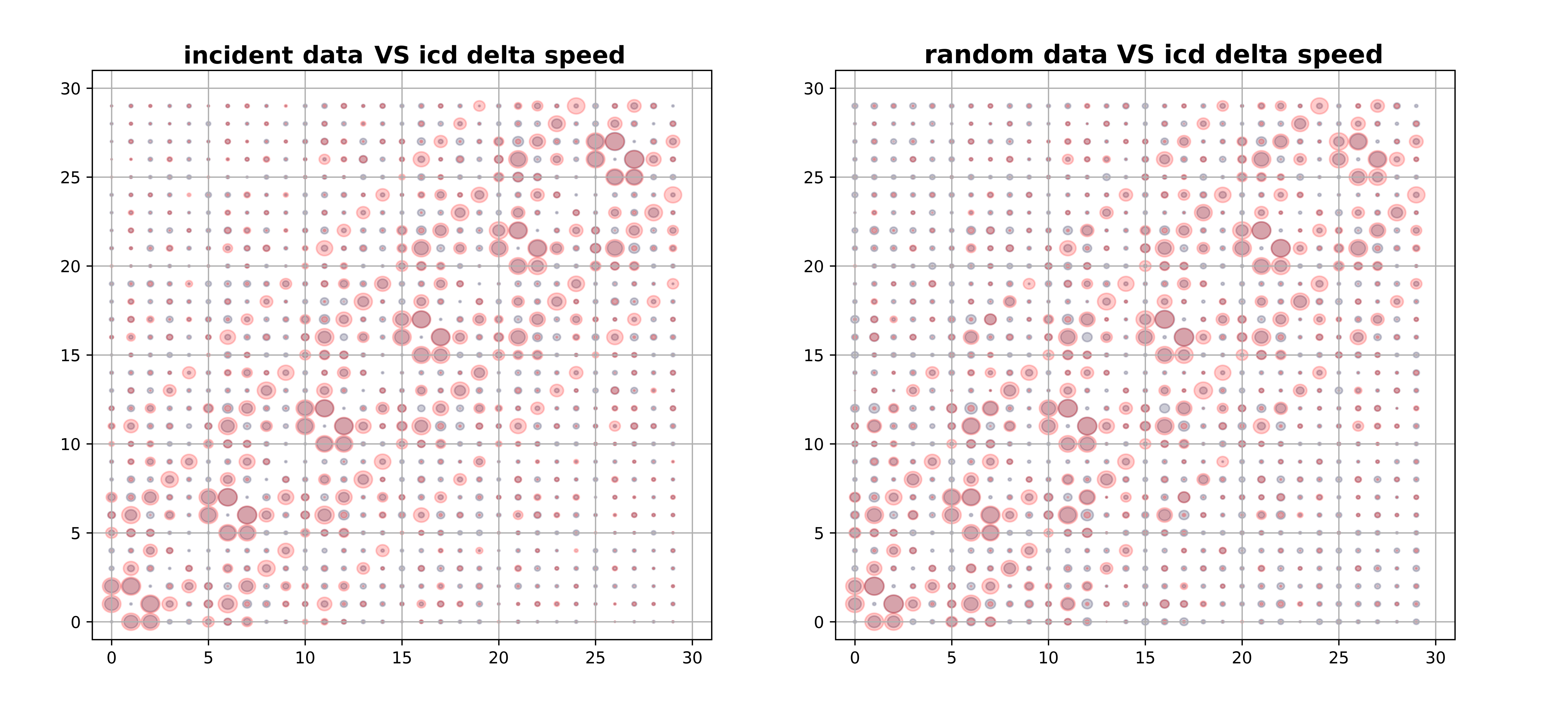}
\caption{ICD result vs data correlation, on the left: accident data, on the right: random data. Labels represent the variables defined in Equation (\ref{causVar}). Red circles indicate results deduced from ICD. Gray circles indicate the Pearson correlation coefficients between variables mined from incident data (left) and random data (right).}
\label{causmx}
\end{figure}

\textbf{Correlation between variables} We apply the sampling algorithm mentioned in Appendix \ref{causalex} to the random data and the accident data, and adopt Pearson correlation coefficient \cite{freedman2007r} to calculate the correlation between the different variables. The result can be plotted in Figure \ref{causmx}, in which the size of the circle indicates the strength of its relations. 

Four phenomena can be observed in Figure \ref{causmx}: (1) The relation obtained by causal discovery holds for both random data and accident data, indicating that the speed changes at one station do have an effect on neighboring stations and that this law is universal, without the need to include an additional judgment about whether an accident has occurred; (2) The causality is mainly reflected in spatial and temporal space, there is no clear spatiotemporal intertwined causality; (3) There is another interesting effect that the station $v_i$, its out-neighbor $\mathbb{O}_1(i)$ and in-neighbor $\mathbb{I}_1(i)$ are causally related to each other; (4) For relations in the temporal dimension, variables are more connected to themselves from $t$ to $t+1$ and $t+2$.

\textbf{Case analysis}  In the real world there may be many reasons behind speed changes, such as sudden traffic jams, weather phenomena, and car accidents. In light of these unpredictable external factors, and to ensure that the results and conclusions on causality seen before are founded in reality, it is essential to plot the speed of a node and its neighbors over time and analyze its behaviors. 

To do so, a dataset of notable events is created. These events are characterized by a sudden and lasting spike in speed variation, positive or negative. Due to the large scale of the PEMS-BAY dataset, it is necessary to sample random mini-batches and calculate the weighted speed change in the station and its neighbors. Sorting through these average velocity fluctuations, the most influential ones are selected, i.e. highest and lowest peaks.
\begin{figure}[ht]
\subfloat[Event at station 400714]{
  \includegraphics[width=0.31\linewidth]{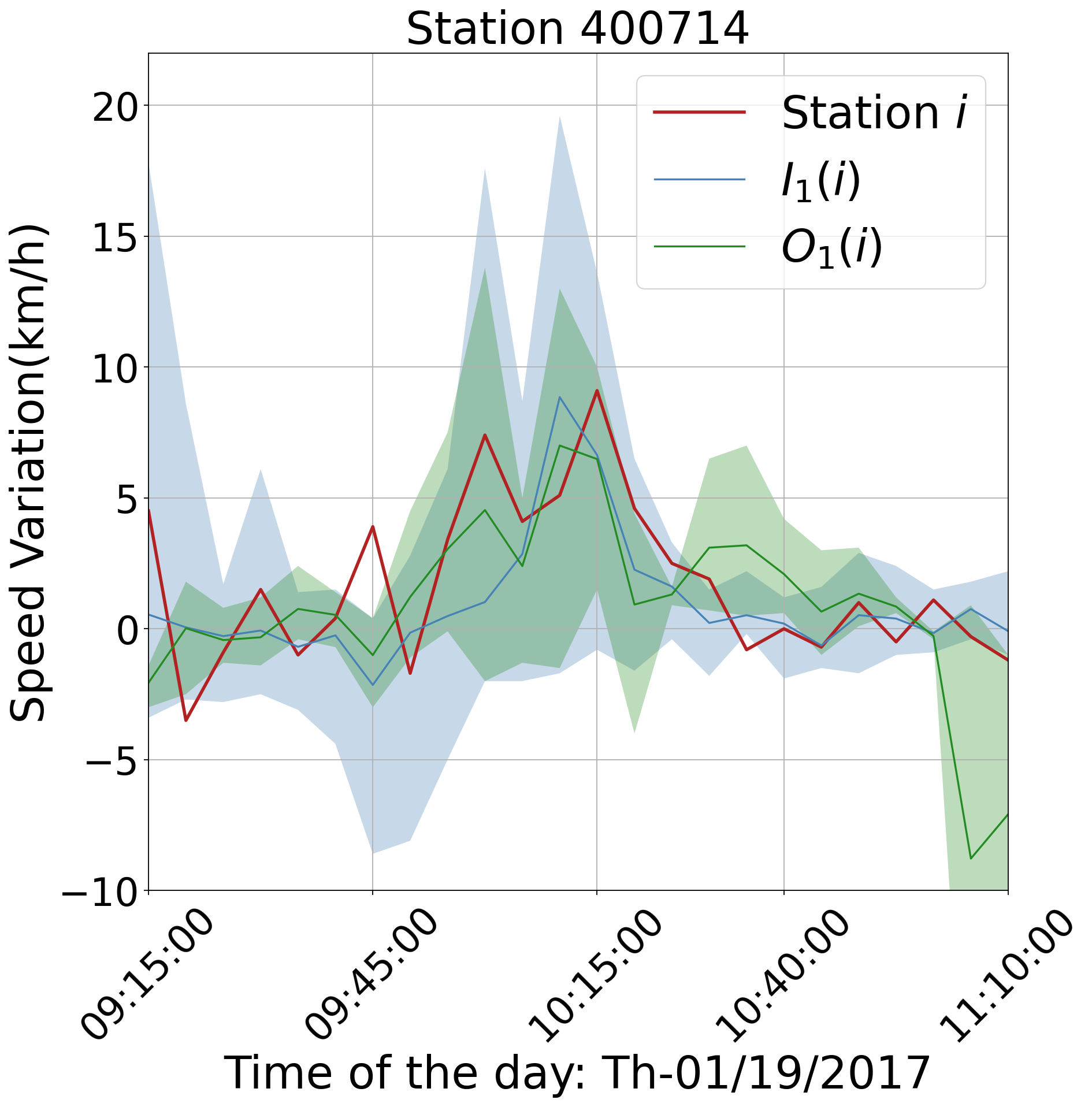}}
  \label{i0i1o1a}
\subfloat[Event at station 407185]{
  \includegraphics[width=0.31\linewidth]{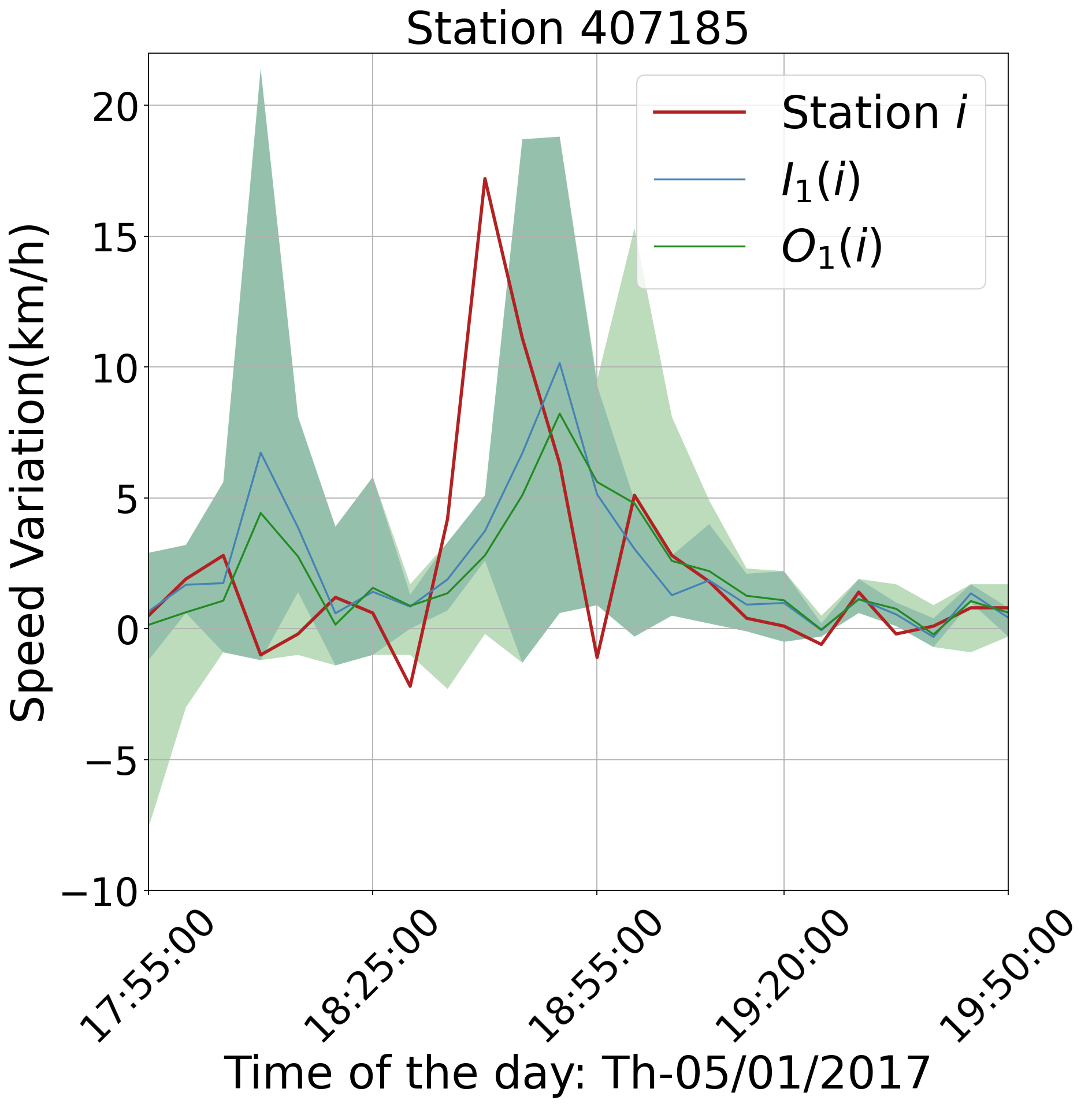}}
  \label{i0i1o1c}
\subfloat[Event at station 400209]{
  \includegraphics[width=0.31\linewidth]{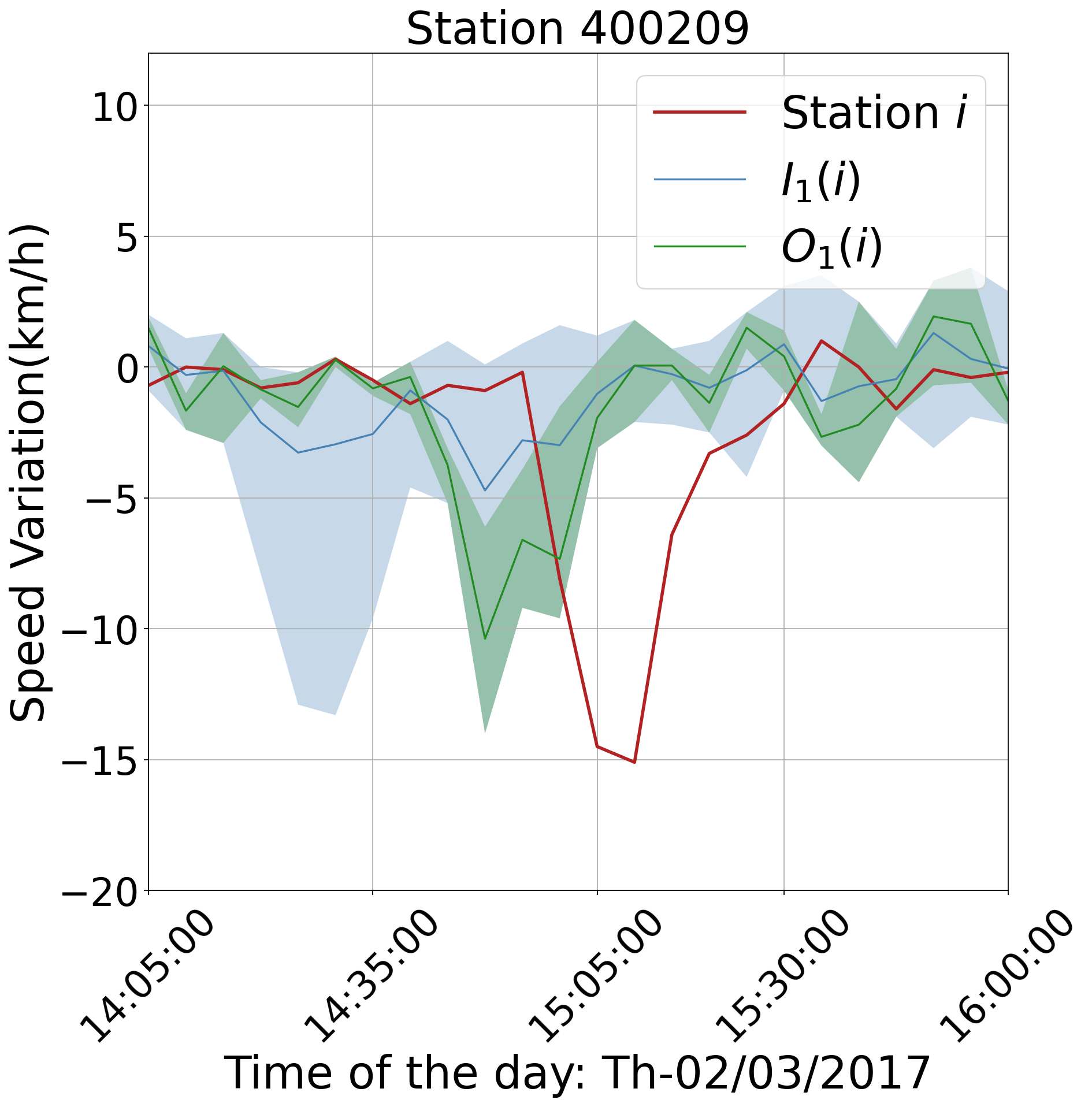}}
  \label{i0i1o1b}
\caption{Notable events displaying the correlation between the station and its neighbors. Curves $I_1(i)$ and $O_1(i)$ denote the weighted sum formulated in Equation (\ref{curve}) and shadow gives the speed variation interval of the first-order in-neighbors/out-neighbors.}
\label{i0i1o1}
\end{figure}

According to these notable events seen in Figure \ref{i0i1o1}, we can confirm the mutual impacts of speed changes between a station, its in-neighbors, and its out-neighbors. In most cases, these three entities share a similar pattern for speed variation, as shown in Figures \ref{i0i1o1a} and \ref{i0i1o1c}. We also observe a delayed impact in some cases, as shown in Figure \ref{i0i1o1b}, where the speed variation is spread from its out-neighbors.

\section{Notation}
\label{notation}
\begin{table}[ht]
\caption{Core concept definitions}
\label{sample-table}
\begin{center}
\resizebox{\linewidth}{22mm}{
\begin{tabular}{ll}
\hline
\multicolumn{1}{c}{\bf Concept}  &\multicolumn{1}{c}{\bf Description}
\\ \hline \\
$\mathcal{G}(\mathcal{V}, \mathcal{E})$  &Graph with node set $\mathcal{V}$ and edge set $\mathcal{E}$, $|\mathcal{V}|=N$\\
$v_i$ &The $i$-th node \\
$\mathbf{W}$ &Adjacency matrix of graph, $\mathbf{W}\in\mathbb{R}^{N\times N}$\\
$\mathbf{D}_I,\mathbf{D}_O$  &In-degree/out-degree matrix \\
$\mathbf{1}$ &All one vector\\
$\mathbf{X}, \hat{\mathbf{X}}$ &Graph signal and the predicted graph signal\\
$\mathbf{I}$ &Identity matrix \\
$\mathbb{I}_1(i), \mathbb{O}_1(i)$  &Set of first-order in/out-neighbors of $i$-th node \\
$\mathbb{I}_2(i), \mathbb{O}_2(i)$  &Set of second-order in/out-neighbors of $i$-th node \\
$\mathbf{I}_1, \mathbf{O}_1$ &Integrated embedding of first-order in/out-neighbors \\
$\mathbf{I}_2, \mathbf{O}_2$ &Integrated embedding of second-order in/out-neighbors \\
$\text{diag}(\mathbf{A})$ &Column vector containing diagonal elements of $\mathbf{A}$ ($\mathbf{A}$ is a matrix)\\
$\text{diag}(A)$ &Diagonal matrix whose entries are elements of $A$ ($A$ is a vector)\\
\hline
\end{tabular}}
\end{center}
\end{table}

\section{Data visualization}\label{distributiondata}
Below are the data distributions for PEMS-BAY dataset.
\begin{figure}[h]
\subfloat[Speed variation distribution]{
  \includegraphics[width=0.5\linewidth]{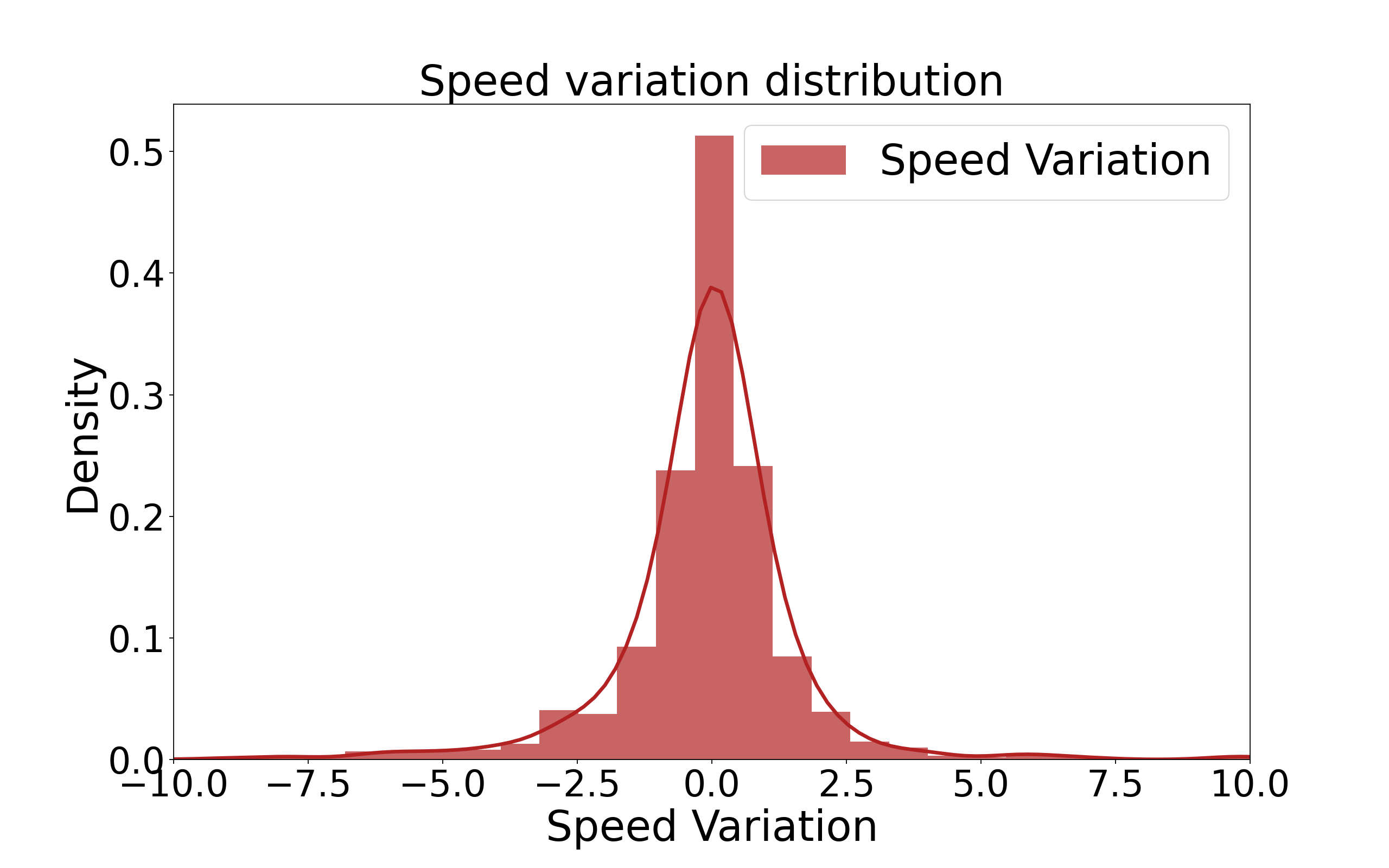}}
\subfloat[Speed distribution]{
  \includegraphics[width=0.5\linewidth]{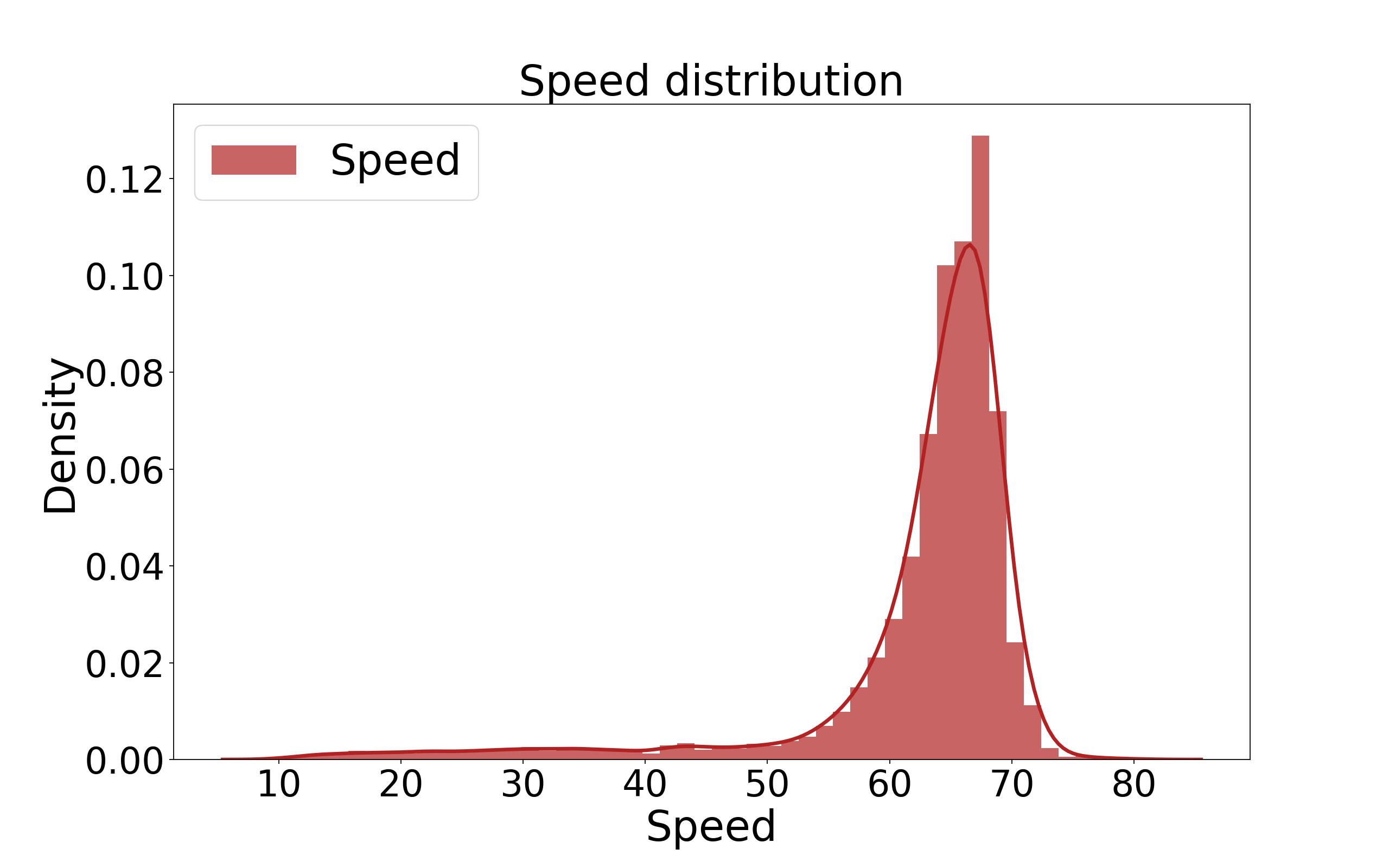}}
\caption{PEMS-BAY data distribution.}
\label{distribution}
\end{figure}

\section{Attention score analysis}
\label{attentionScore}
In this section we focus on the 60min ahead prediction task on station 717477 of METR-LA at 11:15:00. As seen in Figure \ref{subfiga}, there is a performance gain for GT-CausIn in comparison to GT-BadCausIn model. This is due to a structural difference seen in \ref{ablasection} which includes neighbor-level information in the causal insight layer. 
% \begin{figure}[!t]
% \centering
% \includegraphics[width=.5\linewidth]{pictures/prettiest/60minSpeedEval.png}
% \caption{Ground truth of station 71447/neighbors, 60min ahead prediction speed of station 71447.}
% \label{apF1}
% \end{figure}

To look into what is happening inside the causal insight layer, we take out the score matrices. We revise that the attention scores are calculated by processing the score matrix on a softmax layer \cite{attention}, which guarantees that the sum of each row equals one. Given input $\mathbf{X}$, the attention layer is formulated as below.

\begin{gather}
     \mathbf{Q} = \mathbf{X}\mathbf{W}_q, \mathbf{K} = \mathbf{X}\mathbf{W}_k, \mathbf{V} = \mathbf{X}\mathbf{W}_v\\
     \text{attention} = \text{softmax} \frac{(\mathbf{Q}\mathbf{K}^T)}{\sqrt{d_k}}\mathbf{V}
\end{gather}
where $\mathbf{W}_q, \mathbf{W}_k$ and $\mathbf{W}_v$ are trainable weights, $\frac{1}{\sqrt{d_k}}$ is the scalar factor, $\mathbf{Q}$ is query matrix, $\mathbf{K}$ is the key matrix and $\mathbf{V}$ is the value matrix. The attention score is $\mathcal{S}=\frac{(\mathbf{Q}\mathbf{K}^T)}{\sqrt{d_k}}$, and its element $S_{i,j}$ weights the attention $\mathbf{X}_i$ pays to $\mathbf{X}_j$.

\begin{figure}[!t]
\centering
\includegraphics[width=\linewidth]{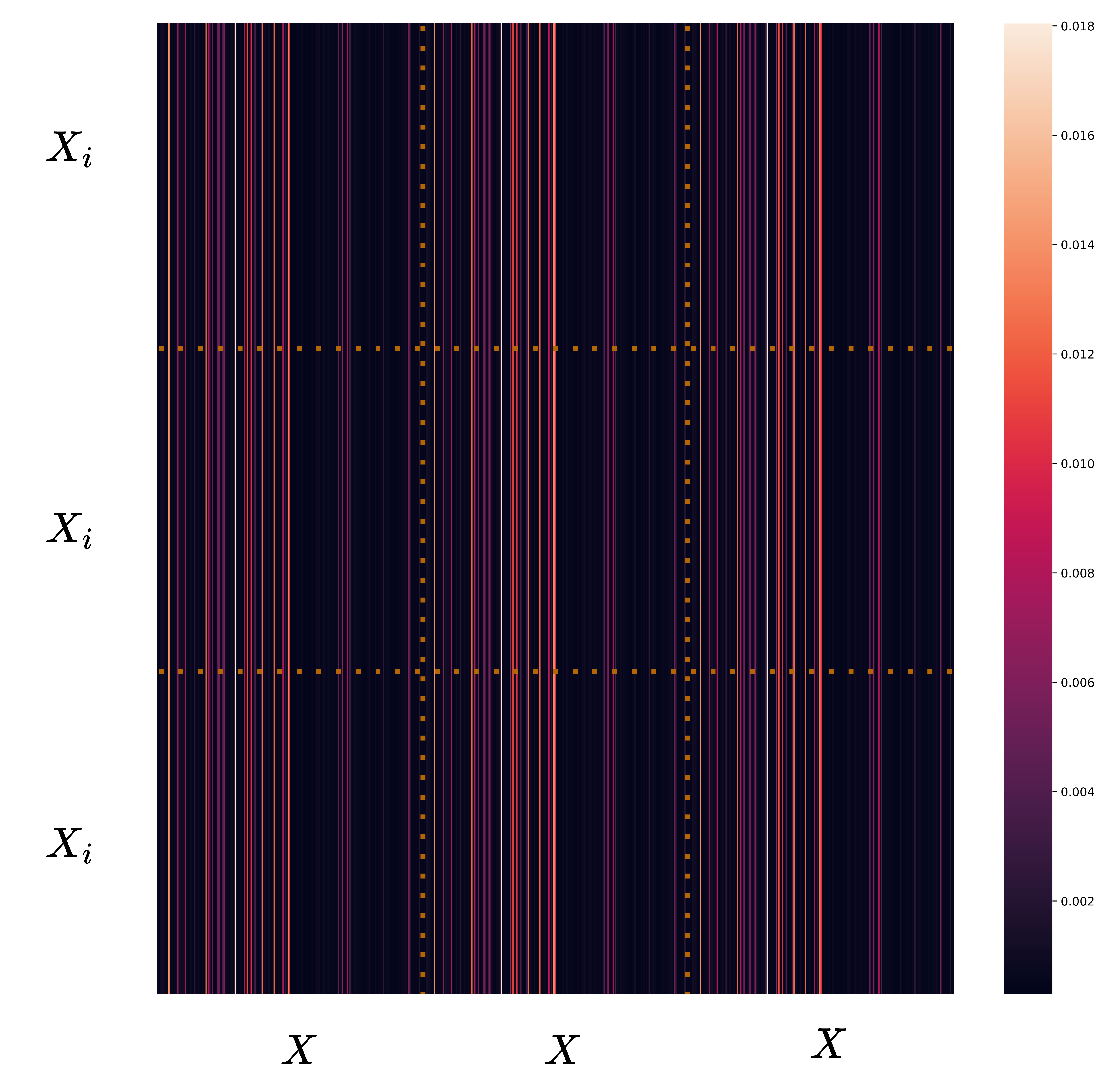}
 \caption{Heat map of attention score in causal insight layer for GT-BadCausIn}
 \label{fig6a}
\end{figure}

\begin{figure}[!t]
\centering
 \includegraphics[width=\linewidth]{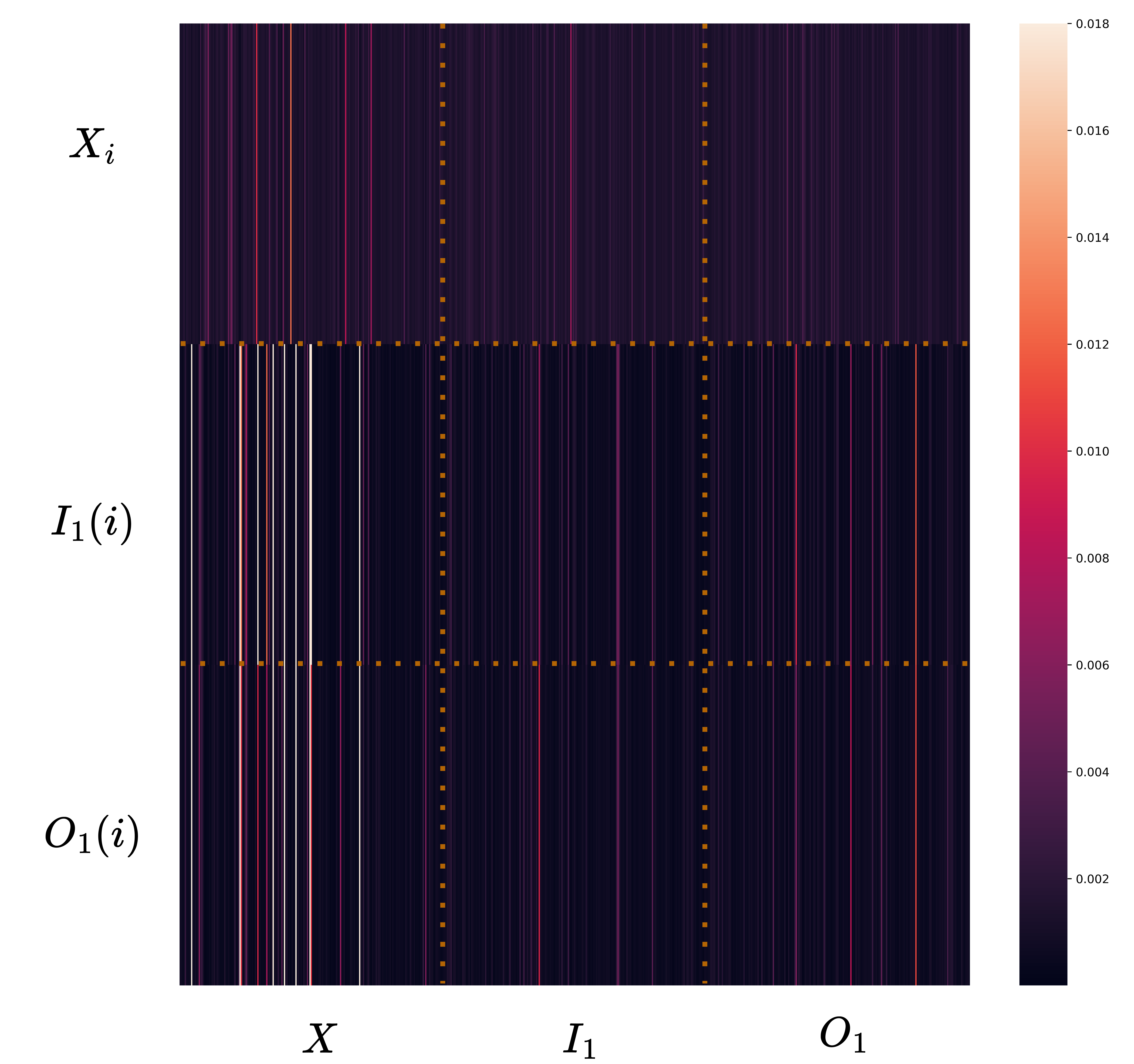}
 \caption{Heat map of attention score in causal insight layer for GT-CausIn}
  \label{fig6b}
\end{figure}

In our case, the attention input is $\mathcal{S} = \mathbf{X}\bigoplus\mathbf{I}_1\bigoplus\mathbf{O}_1$ for GT-CausIn and $\mathcal{S} = \mathbf{X}\bigoplus\mathbf{X}\bigoplus\mathbf{X}$ for GT-BadCausIn, as indicated in Equation (\ref{ab1}) for GT-CausIn and Equation (\ref{abbad}) for GT-BadCausIn. We take rows corresponding to the station $v_i$ to focus on how it is impacted. Afterward, we subdivide the columns into three parts, each part contains tokens of all graph nodes from different perspectives, as noted by the labels in the bottom.

With these tokens in hand, the attention scores are visualized with heat maps. Figure \ref{fig6a} shows attention scores inside the GT-BadCausIn model, that includes solely $\mathbf{X}$ tokens. In contrast, in Figure \ref{fig6b}, the attention scores inside GT-CausIn are not constrained to $\mathbf{X}$, but also including $\mathbf{I}_1$ and $\mathbf{O}_1$ tokens. Finally the performance difference seen in Figure \ref{subfiga} can be explained by different attention scores in the Figures \ref{fig6a} and \ref{fig6b}. In the first case, the attention is mainly distributed on some nodes. In the second case, the attention has a global view on all the nodes thanks to neighbor-level tokens, e.g., $\mathbf{I}_1, \mathbf{O}_1$.

\section{Metrics}
\label{metrics}
Note $\mathrm{x}=\{x_1, \cdots, x_n\} $ the ground truth, $\hat{\mathrm{x}}=\{\hat{x}_1, \cdots, \hat{x}_n\} $ predicted value, and $\Omega$ indices of observed samples, the metrics are defined as follows.
\begin{align}
    \text{MAE}(\mathrm{x}, \hat{\mathrm{x}})&=\frac{1}{|\bf\Omega|}\sum_{i\in{\bf\Omega}}|x_i-\hat{x}_i|\\
    \text{RMSE}(\mathrm{x}, \hat{\mathrm{x}})&=\sqrt{\frac{1}{|\bf\Omega|}\sum_{i\in{\bf\Omega}}|(x_i-\hat{x}_i)^2}\\
    \text{MAPE}(\mathrm{x}, \hat{\mathrm{x}})&=\frac{1}{|\bf\Omega|}\sum_{i\in{\bf\Omega}}|\frac{x_i-\hat{x}_i}{x_i}|
\end{align}

\section{Detailed experimental settings}
\label{exsetting}

For all models, we stack $L=4$ GT blocks (except ablation study on $L$) and the output dimension of each block is 8. In each graph diffusion layer, the maximum number of steps of random walks is 3. The kernel size is 3 for all TCN layers. We use a step learning rate with $\gamma=0.5$ for all experiments and adopt a grid search to find the best initial learning rate for each experiment, as shown in Table \ref{lrsetting}.

\begin{table}[ht]
\caption{Learning rate setting}
\label{lrsetting}
\begin{center}
\begin{tabular}{lll}
\hline
\multicolumn{1}{c}{\bf Data} &\multicolumn{1}{c}{\bf Model}  &\multicolumn{1}{c}{\bf Initial lr\textbackslash start step\textbackslash step size}
\\ \hline \\
\multirow{3}{*}{PEMS-BAY} 
&GT-NoCausIn &0.002\textbackslash 50\textbackslash 20\\
&GT-CausIn &0.006\textbackslash 50\textbackslash 20 \\
&GT-BadCausIn &0.007\textbackslash 50\textbackslash 20\\
\hline \\
\multirow{6}{*}{METR-LA} 
&GT-NoCausIn &0.008\textbackslash180\textbackslash50\\
&GT-CausIn &0.004\textbackslash180\textbackslash50 \\
&GT-BadCausIn &0.004\textbackslash180\textbackslash50\\
&GT-CausIn$(L=3)$ &0.005\textbackslash180\textbackslash 50 \\
&GT-CausIn$(L=5)$ &0.006\textbackslash180\textbackslash 50 \\
&GT-CausIn$(L=6)$ &0.004\textbackslash180\textbackslash 50 \\
\hline \\
\end{tabular} 
\end{center}
\end{table}

\end{document}